\documentclass[10pt,journal,compsoc]{IEEEtran}

\ifCLASSOPTIONcompsoc
  \usepackage[nocompress]{cite}
\else
  \usepackage{cite}
\fi

\usepackage{cite}                      %
\usepackage{tabu}                      %
\usepackage{booktabs}                  %
\usepackage{amsmath}
\usepackage{amsfonts}
\usepackage{colortbl}

\ifCLASSINFOpdf
   \usepackage[pdftex]{graphicx}
\else
\fi

\usepackage{url}

\usepackage{xcolor}
\definecolor{mygreen}{RGB}{20, 190, 10}
\definecolor{myorange}{RGB}{250, 86, 36}
\definecolor{myblue}{RGB}{100, 100, 255}
\newcommand{\ysq}[1]{\textcolor{black}{#1}}
\newcommand{\frysq}[1]{\textcolor{black}{#1}}
\newcommand{\srysq}[1]{\textcolor{black}{#1}}
\newcommand{\trysq}[1]{\textcolor{black}{#1}}
\definecolor{tabred}{RGB}{243, 93, 79}
\definecolor{taborange}{RGB}{242, 144, 105}
\definecolor{tabyellow}{RGB}{190, 165, 81}
\definecolor{tabgreen}{RGB}{167, 158, 129}
\definecolor{LightGray}{RGB}{215,215,215}
\definecolor{LightLightGray}{RGB}{234,234,234}

\newcommand{\Tref}[1]{Table~\ref{#1}}

\newcommand{\Fref}[1]{Fig.~\ref{#1}}
\newcommand{\Sref}[1]{Section~\ref{#1}}

\begin{document}

\title{3D Question Answering}

\author{Shuquan Ye\textsuperscript{1} \quad Dongdong Chen\textsuperscript{2} \quad Songfang Han\textsuperscript{3} \quad Jing Liao\textsuperscript{1}\thanks{Jing Liao is the corresponding author.}\\
\textsuperscript{1} City University of Hong Kong
\quad \textsuperscript{2} Microsoft Cloud AI \quad
\textsuperscript{3} University of California San Diego\\
{\tt\small shuquanye2-c@my.cityu.edu.hk, cddlyf@gmail.com, s5han@eng.ucsd.edu, jingliao@cityu.edu.hk}
}

\markboth{Journal of \LaTeX\ Class Files,~Vol.~14, No.~8, August~2015}%
{Shell \MakeLowercase{\textit{et al.}}: Bare Advanced Demo of IEEEtran.cls for IEEE Computer Society Journals}

\IEEEtitleabstractindextext{%
\begin{abstract}
  Visual \trysq{q}uestion \trysq{a}nswering (VQA) has \trysq{experienced} tremendous progress in recent years. However, most efforts \trysq{have} only focus\trysq{ed} on 2D image question\trysq{-}answering tasks. 
  \frysq{In this paper, we extend VQA to its 3D counterpart, 3D \trysq{q}uestion \trysq{a}nswering (3DQA), which can facilitate \trysq{a} \srysq{machine}'s perception of 3D real-world scenarios.}
  \trysq{Unlike 2D} image VQA, 3DQA takes the color point cloud as input and requires both appearance and 3D \trysq{geometrical} comprehension to answer the 3D-related questions. To this end, we propose a novel transformer-based 3DQA framework \textbf{``3DQA-TR"}, which consists of 
  two encoders \trysq{to} exploit the appearance and geometry information, respectively. \trysq{Finally, t}he multi-modal information \trysq{about the} appearance, geometry, and linguistic question can attend to each other via a 3D-\trysq{l}inguistic Bert to predict the target answers. To verify the effectiveness of our proposed 3DQA framework, we further develop the first 3DQA dataset \textbf{``ScanQA"}, which builds on the ScanNet dataset and contains over 10K question-answer pairs for $806$ scenes. To the best of our knowledge, ScanQA is the first large-scale dataset with natural-language questions and free-form answers in 3D environments that is \textbf{fully human-annotated}. We also use several visualizations and experiments to investigate the astonishing diversity of \trysq{the} collected questions and the significant differences \srysq{between} this task from 2D VQA and 3D captioning. Extensive experiments on this dataset demonstrate the obvious superiority of our proposed 3DQA framework over state-of-the-art VQA frameworks and the effectiveness of our major designs. Our code and dataset will be made publicly available to facilitate research in this direction. The code and data are available at \textcolor{blue}{\url{http://shuquanye.com/3DQA_website/}}.
\end{abstract}
}

\maketitle

\IEEEdisplaynontitleabstractindextext

\IEEEpeerreviewmaketitle

\ifCLASSOPTIONcompsoc
\IEEEraisesectionheading{\section{Introduction}\label{sec:intro}}
\else
\section{Introduction}
\label{sec:intro}
\fi

\begin{figure*}[t] 
  \centering
  \includegraphics[width=\linewidth]{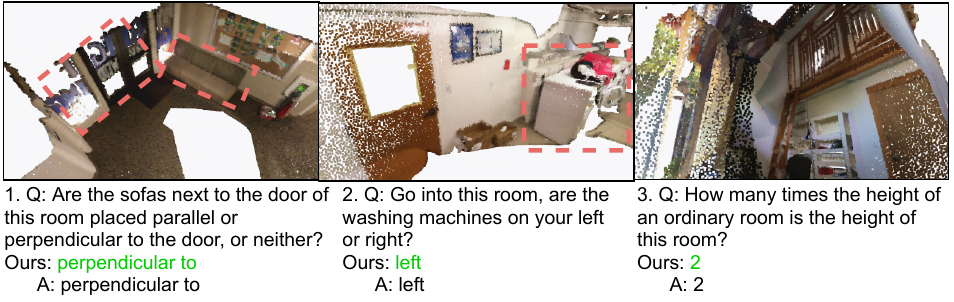}
  \caption{Illustration of typical data samples of \textbf{natural-language, free-form, open-ended} questions collected for 3D scenes in our ScanQA dataset \trysq{with the} answers predicted by our 3DQA framework 3DQA-TR. Objects related to questions are highlighted in the corresponding rendering picture of 3D scans for better visualization. } 
  \label{teaser}
\end{figure*}

\IEEEPARstart{I}{n} recent years, we \trysq{have witnessed} tremendous \trysq{artificial intelligence (AI) progress} \srysq{in} vision and language understanding. Among them, \trysq{v}isual \trysq{q}uestion \trysq{a}nswering (VQA)~\cite{antol2015vqa,vqav2,gqa,clevr,gqaodd,vqacp}, which finds \trysq{the} correct answers to questions based on understanding images, \trysq{has} attract\trysq{ed} a \trysq{significant amount} of research effort. In this area, a number of datasets with well-defined tasks and evaluation protocols \trysq{have been} introduced and various methods~\cite{antol2015vqa,vqav2,gqa,clevr,gqaodd,vqacp} \trysq{have been} proposed.

While existing works in VQA are restricted \trysq{primarily} to images, we take the first step \trysq{toward} extending it to the 3D \trysq{q}uestion \trysq{a}nswering (3DQA) task, i.e., answering questions given \frysq{a color} point cloud. A well-defined 3DQA task will broaden AI's perception to 3D spatial understanding that mimics real-world scenarios and \trysq{will} benefit a wide range of applications, such as \trysq{robot} interactions in real-world environments, information quer\trysq{ies} in augmented \trysq{and} virtual reality, and linguistic-based navigation of autonomous vehicles.
However, extending existing VQA methods to solve 3DQA is non-trivial. Unlike VQA\trysq{, which} relies on 2D \textit{appearance} information to answer questions, 3DQA has \trysq{a} significantly \trysq{greater need to understand} the 3D \textit{geometry}. For example, answer\trysq{ing} the first question \trysq{shown in} \Fref{teaser} \emph{``Are the sofas next to the door of this room placed parallel or perpendicular to the door, or neither?"} requires understand\trysq{ing} not only the appearance, but also the \textit{geometry} structure of \trysq{the} individual objects, and even \trysq{the} \textit{spatial relationships} among different objects.

To address \trysq{these} challenges, we propose the first \srysq{transformer-based} 3DQA framework \textbf{``3DQA-TR"}. It uses two encoders to extract geometry and appearance information from the point cloud and color point cloud, respectively. Given these appearance and geometry encodings along with question embedding, a 3D-Linguistic Bert (3D-L BERT) performs both intra-modal and inter-modal fusion to predict the target answer.  
Specifically, in the geometry encoder, we not only consider the geometry feature\trysq{s} of individual objects, but also explicitly incorporate the coordinates and scales to the spatial embedding in order to model the spatial relationship between objects. \trysq{Moreover,} to provide rich appearance information \trysq{for the appearance encoder}, we pretrain it on a synthetic dataset tailored for appearance information extraction.

\trysq{In addition to} the framework, we also collect the first 3DQA dataset ``\textbf{ScanQA}". It builds upon the real-world indoor scene dataset ScanNet \cite{dai2017scannet}, which contains 1613 scans from $806$ scenes. \trysq{A}nnotators \trysq{were} free to change the viewpoint \trysq{in the 3DQA dataset collection} and ask different types of questions, such as object appearance, object geometry, spatial relationship\srysq{,} and their comparison. After carefully filtering and cleaning, we finally get \ysq{$10,062$ question-answer pairs.} Along with the answers, \trysq{the} annotators' confidence is also provided. Some \trysq{sample} questions are shown in \Fref{teaser}.

To demonstrate the superiority of our framework, we compare\trysq{d} it with representative VQA methods \trysq{answering} 3DQA questions given images from ScanNet videos. The \trysq{comparatively excellent performance} of our framework demonstrates the necessity of \trysq{including} spatial information in the 3DQA task and the effectiveness of our method in exploiting both geometry and appearance information. \trysq{An e}xtensive ablation \trysq{study} also demonstrate the effectiveness of our designs. %

In summary, our contributions are threefold:

\begin{itemize}
    \item 3DQA task: \frysq{We introduce the novel 3DQA task, which involves both language processing and 3D scene understanding.}\footnote{\srysq{Note that 3DQA is different from multi-view or RGB-D QA, and it directly captures 3D information in point clouds.}}
    \item 3DQA-TR framework: We design a new transformer-based framework 3DQA-TR to solve\trysq{ 3DQA} task. It utilizes one language tokenizer for question embedding and two encoders for extracting the appearance and geometry information, respectively, and then uses a 3D-L BERT to perform multi-modal fusion for question answering. %
    \item \textbf{ScanQA} dataset: We \trysq{took} the lead in building the \textbf{fully human-annotated} 3D question answering dataset, ScanQA, which provides natural, free-form, and open-ended questions and answers in free-perspective 3D scans. %
\end{itemize}

We will make our data collection tools, ScanQA dataset and code public to facilitate future research.

\section{Related Work}
\label{sec:related}

\subsection{VQA}\label{subsec:2D} 
The VQA task, which combines \trysq{the} challenges of both visual and linguistic processing to answer questions about \trysq{a} given images, has attracted intense research efforts. Many different VQA datasets and methods have been proposed in the last few years. The most famous VQA datasets include VQAv1~\cite{antol2015vqa} and the VQAv2~\cite{vqav2} annotated by humans, GQA~\cite{gqa} and CLEVR~\cite{clevr} with synthetic \trysq{questions and answers} from real-world or generated images. Based on them, several splits \trysq{have been} further established to study bias and language prior, \trysq{such as} the GQA-OOD~\cite{gqaodd} dataset of infrequent concepts, and the \srysq{bias-sensitive} VQA-CP~\cite{vqacp} dataset. As for VQA methods, \trysq{advances} in deep learning have brought tremendous success in solving VQA tasks by utilizing multi-view~\cite{8885753,bansal2020visual}, panoramic~\cite{9093452,Yun2021PanoAVQA}\srysq{,} RGB-D~\cite{banerjee2021weakly}, and video~\cite{Yun2021PanoAVQA} to capture the 3D information. However, none of \trysq{the methods} directly captures 3D information like ours\srysq{, by utilizing 3D point clouds}. %
Generally, the VQA model~\cite{lu2016hierarchical,yang2016stacked,Anderson2017up-down,kim2021mlb,singh2018pythia,9552236} consists of three components - an image encoder \trysq{to extract} visual information, a language encoder \trysq{to encode} question\trysq{s}, and a fusion module \trysq{to aggregate} information and \trysq{classify} answer\trysq{s}.
Recently, \trysq{t}ransformers~\cite{vaswani2017attention}, the de-facto standard model for language tasks, \trysq{have} been successfully applied to VQA as well. For example, LXMERT~\cite{tan2019lxmert}, VilBERT~\cite{lu2019vilbert} and VL-BERT~\cite{Su2020VLBERT}, which extend the popular BERT architecture to accept joint representations of image content and natural language, have demonstrated the superiority of transformers in solving VQA tasks.

\trysq{In addition to} standard VQA, there are some extensions of VQA to other sub-areas, such as diagrams and document analysis~\cite{ebrahimi2017figureqa,kafle2018dvqa,Chaudhry2020LEAFQALE,mathew2021docvqa}, video understanding~\cite{lei2018tvqa,lei2019tvqa,zadeh2019social,garcia2020knowit}, multiview and viewpoint selection~\cite{s20082281}, \srysq{knowledge-based} \trysq{question answering}~\cite{wang2017fvqa,marino2019ok,garcia2020knowit}, visual commonsense reasoning (VCR)~\cite{zellers2019recognition}, visual dialog~\cite{das2017visual} and embodied question answering (EQA) with navigation (VLN)~\cite{das2018embodied,anderson2018vision}. However, these extensions are still limited to 2D images and there is no extension of VQA from 2D to 3D, for answering questions about a given 3D scene. Our work takes the first step towards it and proposes a new transformer-based framework for 3D understanding and linguistic processing.

\subsection{3D Vision and Language.}\label{subsec:3DandEQA}
3D scene understanding with point cloud\trysq{s}, such as segmentation~\cite{ye2021learning,li2021rotation, hu2018semantic}, registration~\cite{9736452}, upsampling~\cite{9351772}, denoising~\cite{8730533}, and generation~\cite{han2021exemplarbased} has \trysq{made} great progress in recent years.
\trysq{There} are \trysq{also} some works \trysq{that} further exploring 3D scene understanding through language, such as 3D object localization (e.g., ~\cite{achlioptas2020referit3d,chen2020scanrefer,yuan2021instancerefer,huang2021text,feng2021free,yangSAT2021}), 3D object captioning (e.g., ~\cite{Chen_2021_CVPR,chen2020scanrefer}) and \ysq{relationship grounding (e.g., ~\cite{su20212,goyal2020rel3d})}. Among them, ScanRefer~\cite{chen2020scanrefer} is a representative work for both 3D localization and captioning, but it requires the input of multi-view photos provided by the ScanNet dataset\trysq{. However, this requirement} greatly limits \trysq{its} application because 3D datasets do not always have multi-view photos along with them. In contrast, our method takes \trysq{the} point cloud as input. Moreover, these methods and datasets for 3D localization and captioning are restricted to objects \trysq{in} limited categories, which is different from our free-form and open-ended question and answering task in real scenes. 

\begin{figure*}[hbt!] \centering
  \includegraphics[width=0.44\textwidth]{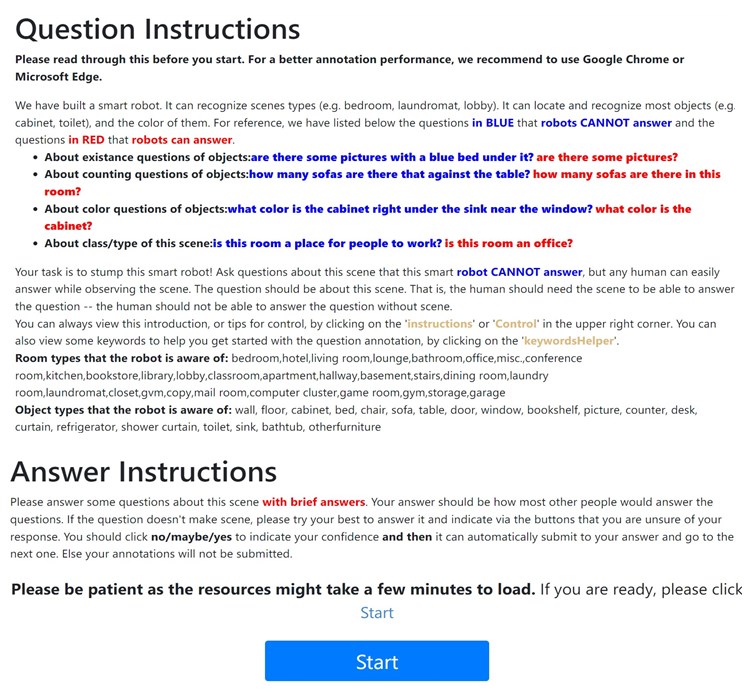}
  \includegraphics[width=0.55\textwidth]{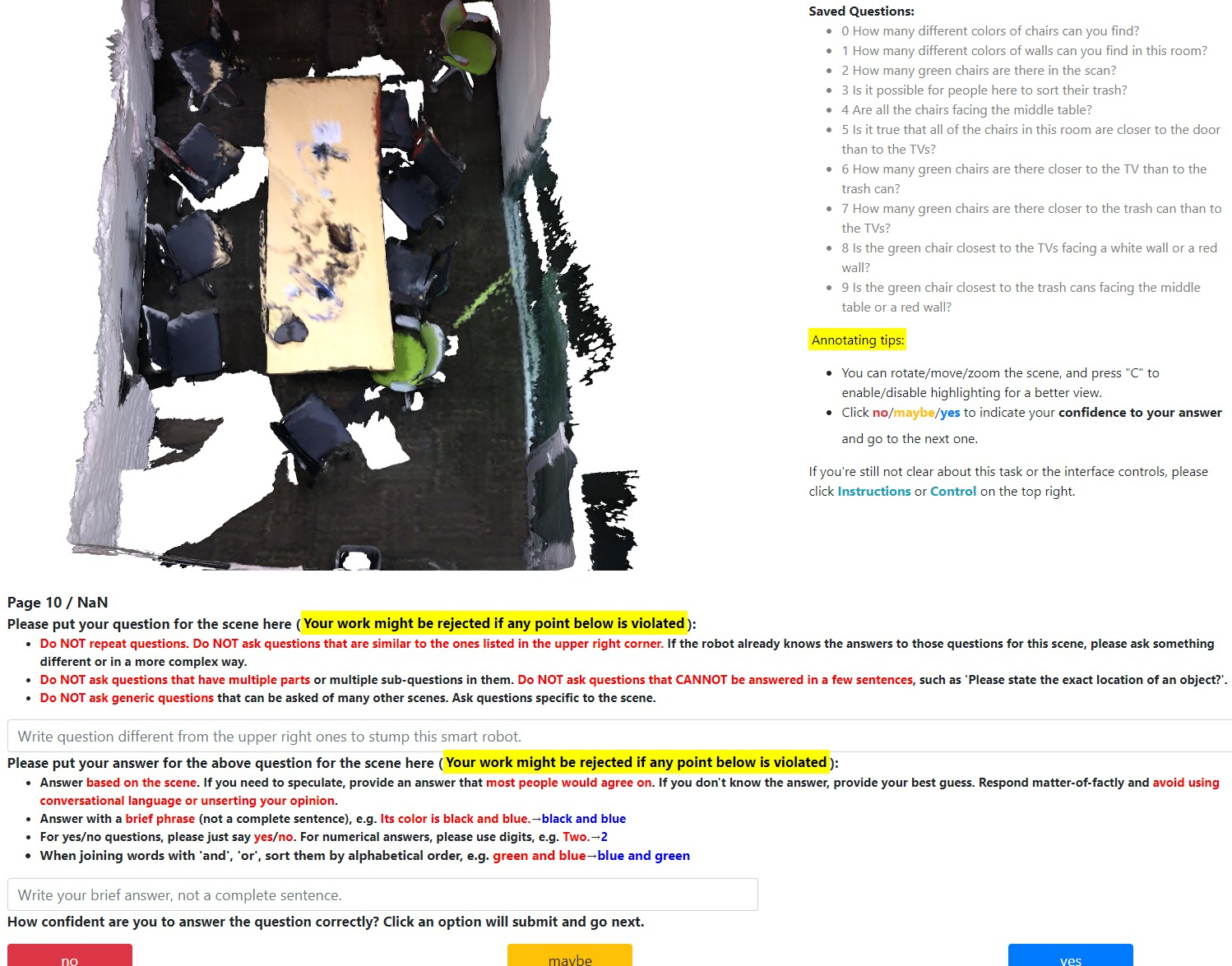}
  \caption{Our \trysq{instruction and annotation pages} for collecting questions and answers in \srysq{the} \trysq{Amazon Mechanical Turk} interface. \trysq{(cropped for better viewing.)}} 
  \label{supp-webqa}
\end{figure*}

In particular, MP3D-EQA~\cite{wijmans2019embodied} proposes a novel navigation task \trysq{using} 2.5D RGB-D frames and language to navigate and answer templated questions in photorealistic environments. First, their task is different from our 3DQA. MP3D-EQA focuses on navigation, so it \trysq{only} supports a few object classes and only three types of template questions (location, color, color\_room) which \trysq{are all} related to navigation. In contrast, our 3DQA questions contain a wider variety of objects, spatial and visual concepts. Moreover, their RGB-D inputs limit viewpoints \trysq{to} observing the 3D scene, while our point cloud input enables free views to meet the higher requirements in scene understanding brought by the free-form and open-ended questions. 
In addition, as their template question follows only a few forms, this task has a very limited problem diversity and low linguistic-processing difficulty. \trysq{Furthermore}, the language model in their approach is not flexible - the text representation is the average of word embeddings obtained by a predefined matrix. %
With the above restrictions tailored for the navigation task, it is difficult to apply MP3D-EQA in our 3DQA task.

\section{3DQA Task and ScanQA Dataset} 
\label{sec:taskanddata}

We introduce the task of 3D Question Answering with natural, free-form, open-ended questions and answers in real-world 3D scenes. Given a 3D scene $\mathcal{S}$ represented by point cloud of $N$ points with color channel $\mathcal{S} \in \mathbb{R}^{N\times6}$ ($xyzrgb$), and a question $\mathcal{Q}$ of $T$ linguistic words \trysq{denoted} as $\{q_1,...,q_t,...,q_T\}$, the goal of 3DQA is to predict the answer $\mathcal{A}$.

In this section, we first present our data collection and quality control \trysq{strategies}. Then we analyze the question and answer statistics of the built dataset, along with the \trysq{analyzing} of \trysq{the} confidence \trysq{levels} and inter-subject agreements, according to which we design the evaluation metric. Based on the analysis, we also clearly show the difference between this 3DQA task and VQA.

\subsection{Dataset Collection}

The ScanQA dataset is built \trysq{on} ScanNet~\cite{dai2017scannet}, a real-world indoor scene dataset of $1,613$ scans from $806$ scenes. 
\trysq{Unlike} the question and answer collection \trysq{for} VQA tasks, our \textit{3D web-based} \trysq{user interface} lets the annotators freely control the viewpoint, camera setting, as well as transparency of each scan. This enables annotators to investigate various aspects of the scenes.

To ensure the \textit{viewpoint variety} of \trysq{the} question and answer annotations, and to encourage sufficient exploration of the scene, we reject those with overlapping camera parameters.
We \trysq{also} design three methods to encourage \textit{interesting and diverse questions}. First, we show the previously asked questions to question annotators and reject duplicated ones. Second, we utilize an online robot and encourage the annotators to stump it. Third, \trysq{we prompt the annotations with} a rich collection of carefully selected interesting questions shown on the instruction pages.
In order to ensure the \textit{quality} of questions and answers, besides the aforementioned online robot checker, syntax check and manual correction are also performed to reduce possible grammatical errors. %
All questions and answers are collected by crowdsourcing on Amazon Mechanical Turk (AMT). %
\frysq{Along} with the answers, the annotator's confidence is also recorded. %

\begin{figure*}[t] 
  \centering
  \includegraphics[width=0.68\textwidth]{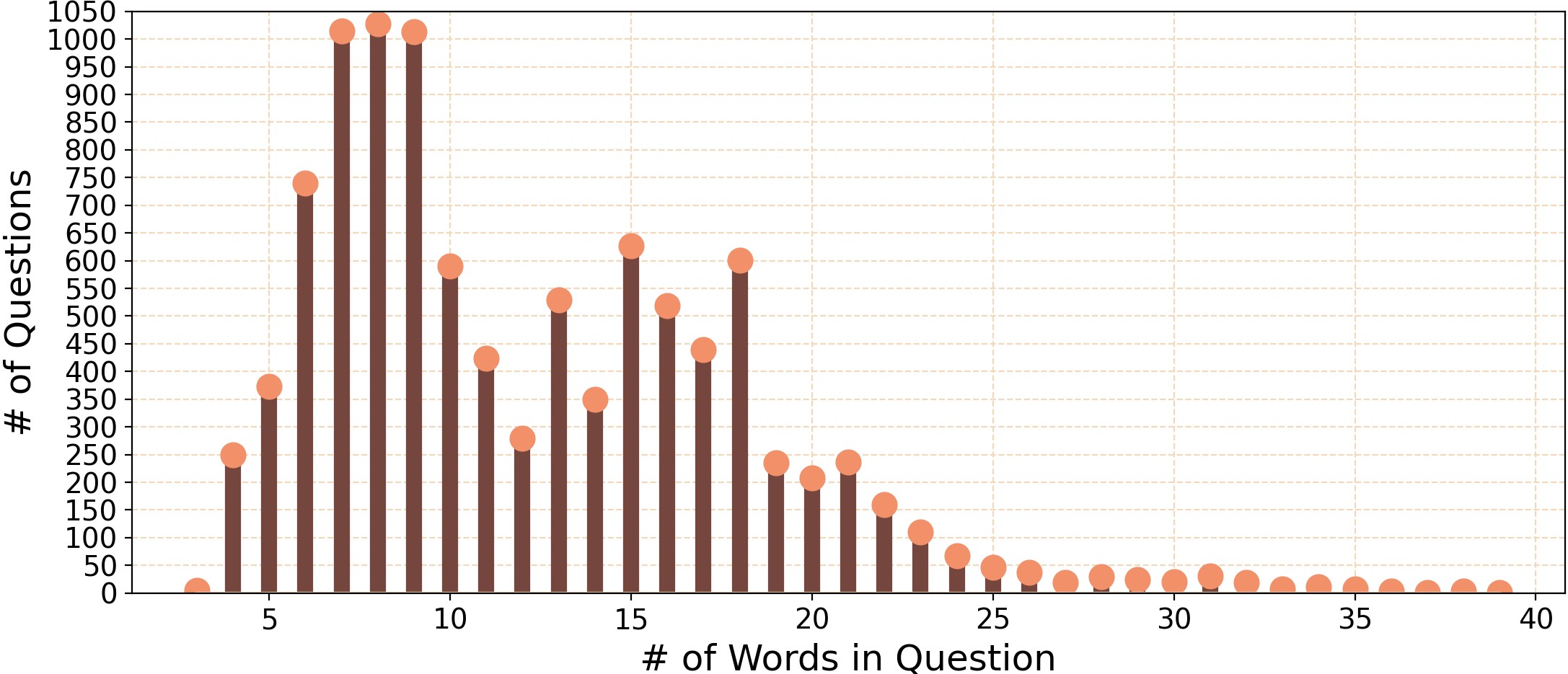}
  \includegraphics[width=0.275\textwidth]{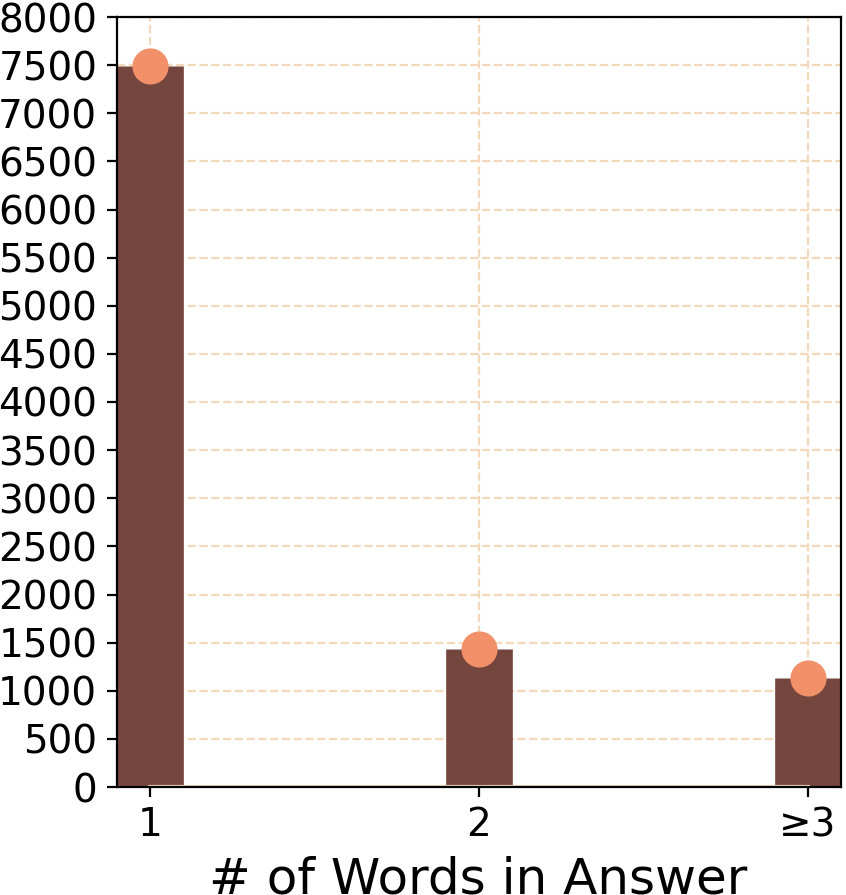}
  \caption{Question and \trysq{answer length} distribution.} 
  \label{questionlength}
\end{figure*}

\ysq{Fig.~\ref{supp-webqa} \trysq{shows} the detailed instruction page and annotation page for collecting questions and answers.
As shown in the instruction page (the left image), to \trysq{improve} the \textit{quality and diversity} of \trysq{the} questions, we claim that annotators \trysq{they} need to stump a smart robot.
We created an online robot checker that would reject simple queries of predefined known object classes and known scene types, such as the existence, counts\srysq{, or} color\trysq{s} of known objects.  These predefined known objects (e.g., wall, floor, \trysq{and} cabinet) and scene classes (e.g., bedroom, hotel, \trysq{and} living room) are given on the instruction page. For example, questions \trysq{such as} `What color is the table there?', `Is this room a kitchen?' will be rejected, and questions \trysq{such as} `Is there a computer in front of the swivel chair closest to the door?' will not. 
\trysq{Specifically}, we define several patterns of easy-to-answer questions and reject questions that fit into these patterns:}
\begin{itemize}
\item Existence Pattern: Is$|$Are there a$|$an$|$any$|$some $<$KnownObjectClass$>$ ...? (with $<8$ words)
\item Count Pattern: How many $<$KnownObjectClass$>$ [is$|$are$|$in in$|$there$|$this$|$the] ...? (with $<9$ words)
\item Color Pattern: What color$|$colour is$|$are the $<$KnownObjectClass$>$ [is$|$are$|$in in$|$there$|$this$|$the] ...? (with $<6$ words)
\item Scene Type Pattern: Is it$|$this [scene$|$scan$|$room] a$|$an $<$KnownSceneClass$>$ ...?
\end{itemize}

\ysq{To promote interesting and diverse questions, we present annotators with previously asked questions and reject duplicated ones, as illustrated in Fig.~\ref{supp-webqa} (right). In addition, we encourage annotators to ask questions specific to one scene rather than generic questions. 
 \trysq{To do this}, we present annotators with scene scans in the form of colored meshes, allowing annotators to freely modify the viewpoint and camera settings via rotation, mov\trysq{ing}, and zoom\trysq{ing}, as well as \trysq{modifying} the transparency of \trysq{the} mesh surfaces. }

\subsection{ScanQA Dataset} \label{subsec:dataanalysis}

In total, the ScanQA dataset provides $10,062$ question-answer pairs (\trysq{an average of} $12.5$ question-answer pairs per scene). \trysq{Further} details can be found in the supplementary materials.

\subsubsection{Question and Answer Statistics}

Fig~\ref{questionlength} \trysq{shows} the length distribution of \trysq{the} natural-language, free-form, open-ended questions \trysq{in} our dataset. It can be seen that the majority of the length of \trysq{the} questions \trysq{have} between $7$ and $18$ words, and the average length of the questions is $12.51$\trysq{which is more than twice as long as} the average question length of $5.1$ of \srysq{the} VQA dataset, one of the most representative datasets in the image VQA task\trysq{s}, reflecting the high quality of our ScanQA dataset and the complexity of the 3DQA task.

In Fig.~\ref{questionlength}, we show the length distribution of the answers of our ScanQA dataset. While \trysq{most} of our answers are \frysq{one word long}, there are still $25.51\%$ of the answers \trysq{consist} of two or more words. It's much more than the VQA dataset which has only ($9.65\%$) answers containing two or more words.

We also note that \trysq{simple answers do} not imply that the questions are straightforward. On the contrary, the questions usually require complex\trysq{, multi-step} reasoning\trysq{. For example, the Yes/No answer to} question 2 in \Fref{fig:showdata} requires recognition of fine-grained class objects (massage chairs), and aggregation \trysq{the} placement information from multiple objects.

\subsubsection{Question Distribution Based on \trysq{the} First Three Words} %

\begin{figure}[t] 
\centering
\includegraphics[width=0.48\textwidth]{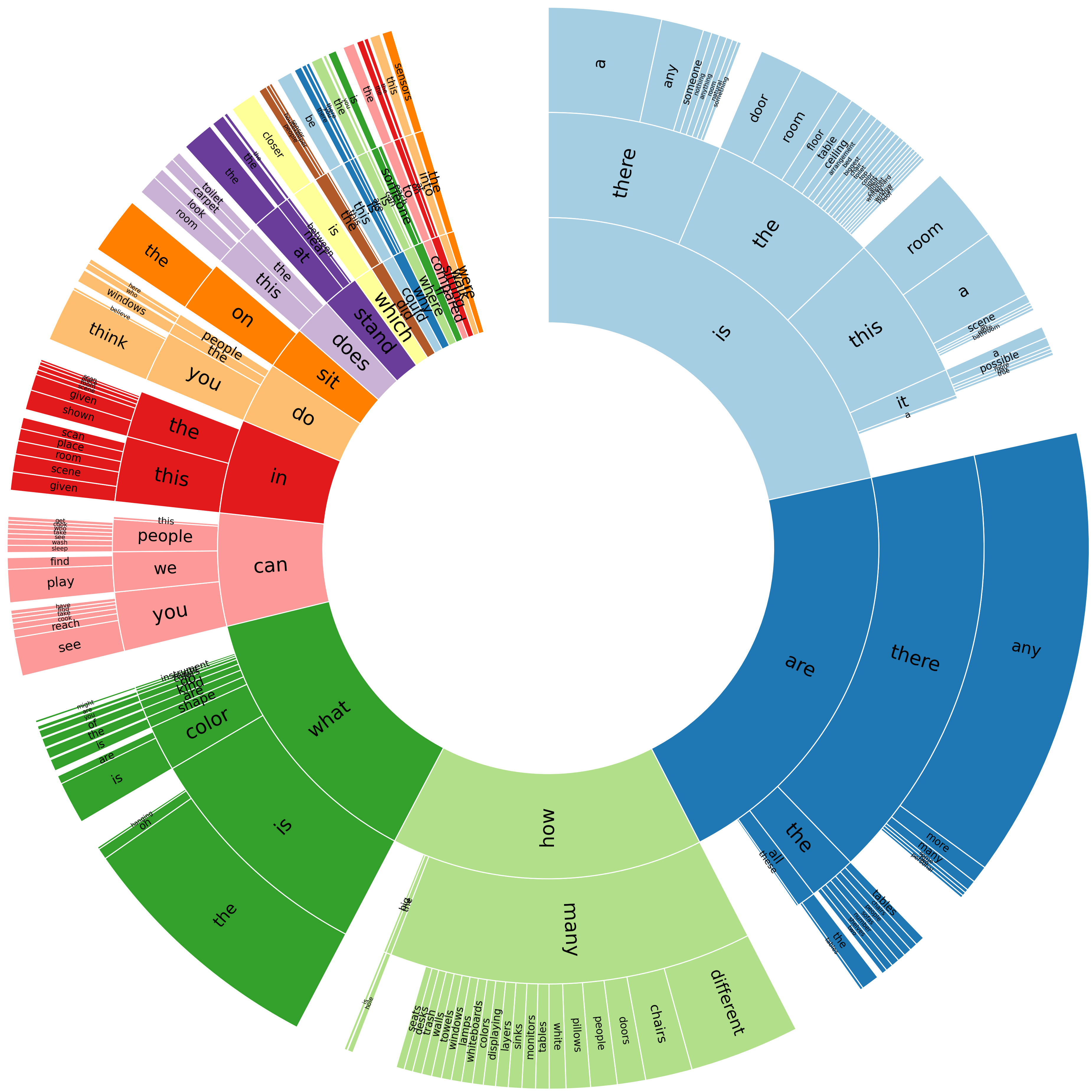}
\caption{Question distribution by their first three Words in ScanQA.}
\label{supp-firstthree}
\end{figure}

In Fig.~\ref{supp-firstthree}, we show the distribution of questions, \trysq{grouped} roughly by their first three words. \trysq{In addition to} questions that require appearance information (e.g., questions \trysq{that} begin with `\trysq{I}s there a ... colored', `\trysq{A}re there any ... colored', `\trysq{H}ow many different colors', `\trysq{W}hat color is/are', `\trysq{I}s there natural light'), there are several types of questions that capture spatial concepts (e.g.,  `\trysq{I}s this room spacious', `\trysq{I}s this a narrow/large/small', `\trysq{I}s this a large/', `\trysq{W}hat is on/hanging/next', `\trysq{W}hat shape', `\trysq{C}an you reach') and placement (e.g., `\trysq{I}s the arrangement'). \trysq{In addition} a certain percentage of the questions shown here necessitate navigation ability (e.g., `stand/sit/standing/sitting') as well as the ability to aggregate from more than two objects (e.g., `\trysq{I}s/are there any', `\trysq{H}ow many different', `\trysq{A}re all', `\trysq{W}hich is closer to').

\subsubsection{Answer Distributions over Different Type of Questions}

\begin{figure}[t] 
\centering
\includegraphics[width=0.485\textwidth]{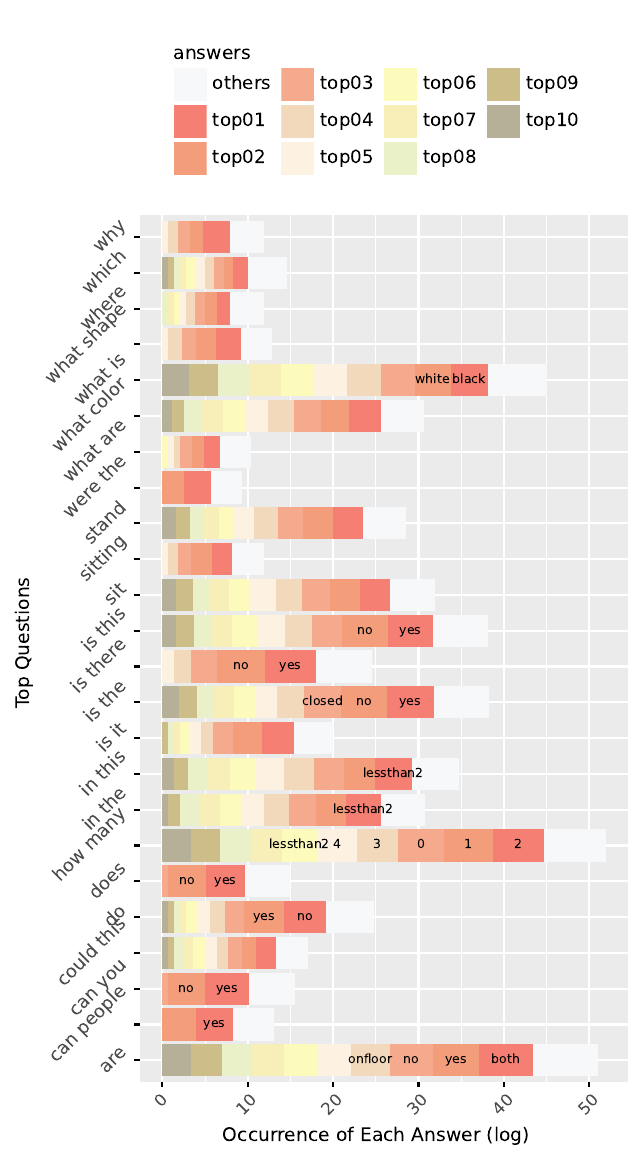}
\caption{Answer occurrence distribution for each of the top question types in ScanQA.}
\label{supp-answer_by_first2}
\end{figure}

\begin{figure}[t] 
\centering
\includegraphics[width=0.46\textwidth]{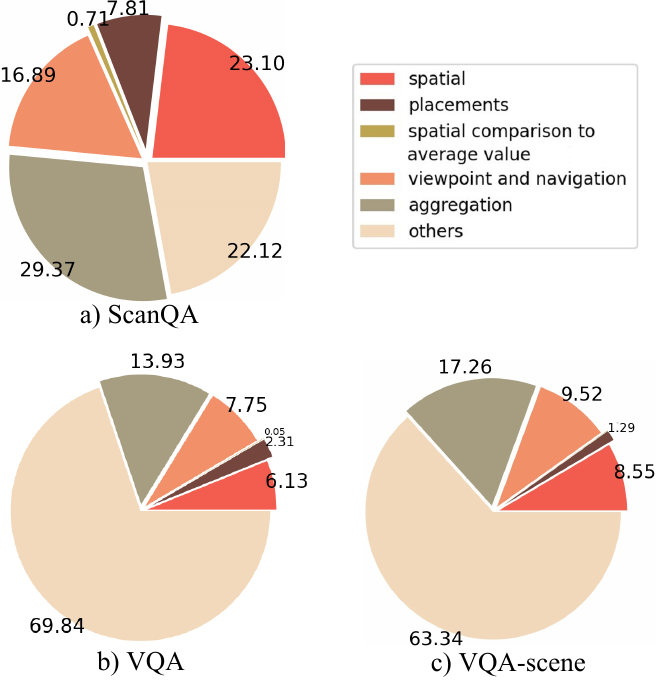}
\caption{Question type comparison in ScanQA, VQA and VQA-scene.}
\label{fig:questiontypes}
\end{figure}

\begin{figure}[t] 
  \centering
  \includegraphics[width=0.485\textwidth]{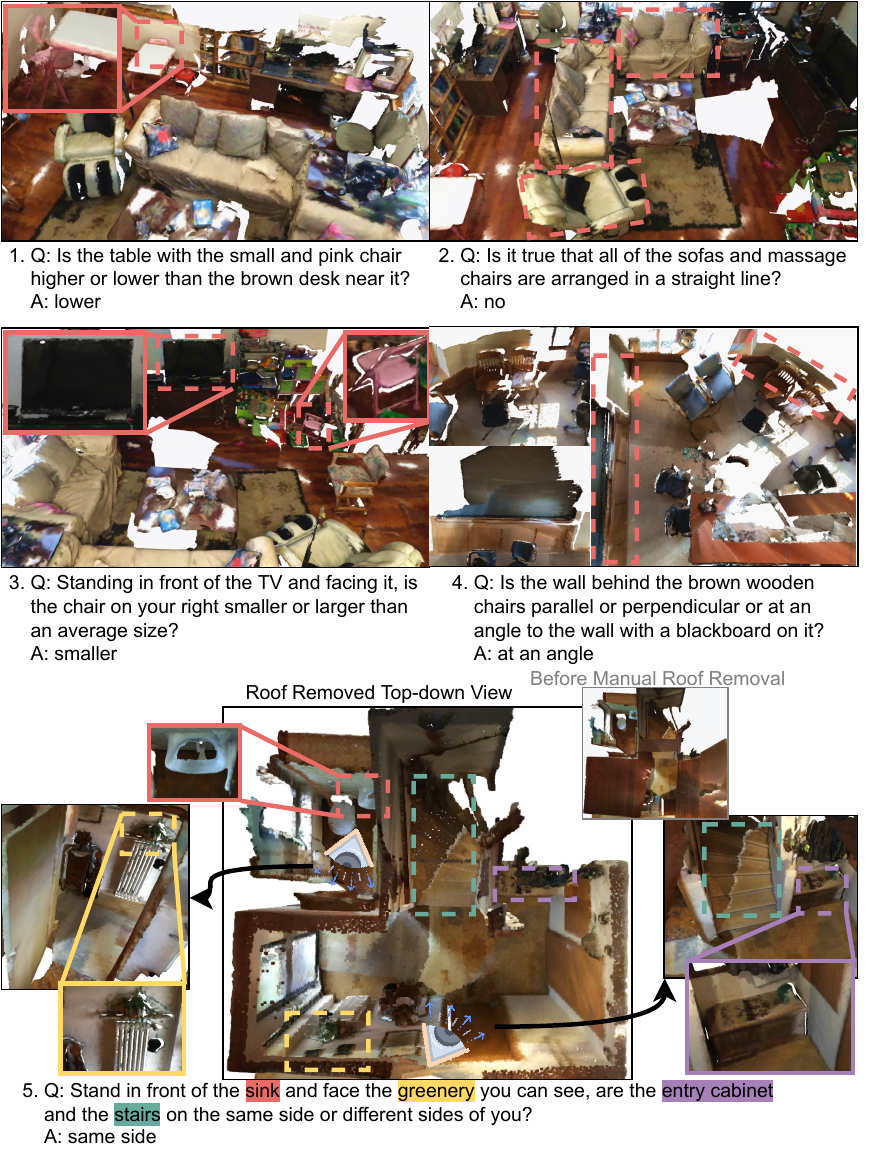}
  \vspace{-0.5em}
  \caption{Illustration of various question types in our ScanQA dataset, which includes a comprehensive variety of appearance and geometry concepts rare in 2D VQA. Question 1 requires \trysq{the} appearance and geometry information of a single object, and \trysq{the} spatial relationships between objects. Question 2 captures the placement information from the aggregation of more than two objects. Question 3 requires navigation capability and spatial comparison \trysq{with} an average value. Question 4 captures complex spatial placement relationships \trysq{that} cannot be drawn from 2D clues – even with the left 2D images \trysq{which} include all \trysq{the} 2D clues in this scene. In question 5, the clues cannot be \trysq{found} within a single view because of occlusions, and the correct answers can be drawn only by observing the 3D scan with two or even more perspectives.}%
  \label{fig:showdata}
\end{figure}

\trysq{Fig.}~\ref{supp-answer_by_first2} \trysq{shows} the distribution of answers across different types of questions, grouped by their first words. The heights of the bars show the number of occurrences for each answer, and the colors indicate different answers.  We observe that only a few types of questions can \trysq{typically be} answered by `yes' or `no'. Thus the majority of question types have rich diverse answers.

\subsubsection{Comparison with VQA Dataset}

Compared to the previous VQA task with 2D images, our 3DQA takes a 3D scene as input and \trysq{is more} concern\trysq{ed with} object geometry and spatial relationship\trysq{s} between objects.  We demonstrate the different characteristics between the VQA dataset and our ScanQA dataset based on \trysq{the} question types. We first categorize \trysq{the} questions into various types and then analyze their statistics \trysq{for} each dataset.

\begin{table}[h]
  \centering
    \caption{Unique nouns rare in VQA and VQA-scene dataset.}
  \label{tab:nounsvsvqa}
  \begin{tabular}{@{}c|c@{}}
  \toprule
  Scenes       & Nouns Rare in VQA \& VQA-scene \\ \midrule
  Apartment        & arrangement, instruments, doorway, bohemian     \\ \midrule
  Bathroom         & privacy, sanitation, towels, toilet, shower     \\ \midrule
  Bedroom    & asymmetrical, bedside, housekeeper, quilt    \\ \midrule
  Classroom & whiteboard, chalkboard, projector, teacher        \\ \midrule
  Conference  & meeting, swivel, monitors, laptops          \\ \midrule
  Laundry & ironing, washing, machines, dryers, washer        \\ \midrule
  LivingRoom & spacious, industrial, scandinavian, neat        \\ \bottomrule
  \end{tabular}
\end{table}

Fig.~\ref{fig:questiontypes} illustrate\trysq{s} the statistics \trysq{by} question types \trysq{for} our ScanQA dataset and \trysq{the} VQA dataset. First, the spatial concept is much more important for 3D scenes, compared to perspective projected and scale agnostic images. Thus, questions about ``\textit{spatial}'' concepts (e.g., scale, angle, position and their comparison), ``\textit{placement}'' (e.g., symmetrical), and ``\textit{spatial comparison to average}'' (e.g., to estimate \trysq{the} size and compare \trysq{it} to an average one) are common in 3DQA. 
Second, compared to the 3D point cloud, 2D images are viewpoint specific, \trysq{arising from} different conventional spatial representations.  For example, we commonly use words like ``front'', or references like ``the closer desk (to me)'' to describe an object in 2D images. However, as these descriptions are \srysq{viewpoint-dependent}, they will cause \trysq{ambiguity} in a 3D scene. Instead, people tend to specify one object in 3D scenes by describing a \textit{viewpoint or doing navigation}. 
Third, compared to a single-view 2D image, a 3D scan reconstructed from thousands of views is less restricted by occlusion. As a result, a 3D scan generally contains much more information than one single image. In order to answer these questions specific to the 3DQA task,  the network should be designed with long-term memory and \trysq{the ability to} ``\trysq{\textit{aggregate}}'' \trysq{information} from more than two concerned objects.
\ysq{We also show some typical data samples from our dataset in \Fref{fig:showdata} to illustrate the various appearance and geometry concepts in 3DQA, which is rare in 2D VQA.}
For example, \trysq{Q}uestion 1 requires both appearance and geometry information \trysq{about} a single object and also spatial relationships with other objects. Question 2 captures the placement information from the aggregation of more than two objects. Question 3 requires the navigation capability to reach \trysq{the} correct viewpoint.
In \trysq{Q}uestion 4, the complex placement relationships cannot be drawn from 2D clues even with the left 2D images that include all \trysq{the} 2D clues in this scene. The correct answer can be drawn only by observing the 3D scan with a perspective that is not humanly possible (right rendering picture). For \trysq{Q}uestion 5, the clues cannot be \trysq{found} within a single view because of occlusions. We show three views for this scene. The middle view is the top-down view of the whole scene in which the ceiling is manually removed for a better view. Before removing \trysq{it}, the greenery and the entry cabinet are occluded by the roof. The view on the left is a view from the sink to the greenery, where both the entry cabinet and the stairs are obscured by the wall. The view on the middle right is a top-down view of the entry cabinet and stairs, where the sink and greenery are occluded by the wall and the roof. To \trysq{obtain} the correct answer with 2D images as input, at least two images are needed with carefully-selected views, while \trysq{selecting} the correct views and obtaining \trysq{the} camera positions from \trysq{the} images or videos are also extremely difficult.

Another way to demonstrate the difference \trysq{between} 3DQA and VQA tasks is to show the top unique nouns that are rare in VQA and VQA-scene datasets. In Table~\ref{tab:nounsvsvqa}, we present the analysis group by different scene types.

\subsubsection{Comparison with 3D Captioning Dataset}

\begin{table*}[t]
\setlength{\tabcolsep}{1.8mm}
\centering
\caption{\ysq{EM of human answers with different input settings, indicating the high quality of our dataset and the importance of 3D scene comprehension. Captions are from ScanRefer~\cite{chen2020scanrefer}.} }
\label{tab:humanaccs_vsCap}
\scalebox{1.0}{
\begin{tabular}{@{}c|ccccccccc@{}}
\toprule
 & \multicolumn{9}{c}{Human Accuracy EM (\%)}  \\
{Input}     & \multicolumn{1}{c|}{All} & Number & Color & Y/N & \multicolumn{1}{c|}{Other} & aggregation & placement & spatial & viewpoint \\ \midrule
{Question+Caption}     & \multicolumn{1}{c|}{\textcolor{black}{41.13}}    &  \textcolor{black}{27.96}   &   \textcolor{black}{33.96}     &  \textcolor{black}{52.13}     &  \multicolumn{1}{c|}{\textcolor{black}{34.47}}      &   \textcolor{black}{ 32.00}          &  \textcolor{black}{40.28}         &  \textcolor{black}{34.50}       &  \textcolor{black}{42.48}         \\
{Question+Scene} & \multicolumn{1}{c|}{\textcolor{black}{83.26}}    & \textcolor{black}{79.57}    &  \textcolor{black}{79.25}      &  \textcolor{black}{94.15}     & \multicolumn{1}{c|}{\textcolor{black}{73.95}}      &  \textcolor{black}{76.50}                   &  \textcolor{black}{81.94}         & \textcolor{black}{82.46}        & \textcolor{black}{82.30}          \\\bottomrule

\end{tabular}}
\end{table*}

In this section, we explain the rationality of our dataset being \textbf{fully human-annotated} rather than template-based generated from 3D captioning in several ways. \trysq{W}e show that our 3DQA task requires richer perception \trysq{than captioning tasks on 3D scenes}, as well as having significant differences in \trysq{the} target setup and dataset distribution.

\trysq{By conducting a human study, w}e prove the richer perception required by 3DQA. Table~\ref{tab:humanaccs_vsCap} compares the performance of human participants who are given both \srysq{questions} and 3D scenes, \trysq{to the performance of} other human participants who are given both \srysq{questions} and all \trysq{the} captions of the corresponding scene. Notably, the human performances with captions are significantly lower than \trysq{when} the people \trysq{are} given 3D scenes, indicating that referring expressions are insufficient to solve 3DQA problems. This demonstrates that the 3D scene perception required in 3DQA is beyond \srysq{what can} be captured by captions and emphasizes the importance of 3D scene comprehension in 3DQA.

The target of our 3DQA also has significant differences \trysq{from} 3D captioning. While 3D captions are restricted to objects of limited classes as we mentioned in \Sref{subsec:3DandEQA}, we assume a challenging problem setup where the questions are free-form, allowing it to capture everything from small and detailed items to fine-grained class objects, and even the entire scene.

We statistically measure the dataset distribution differences in the word distributions \trysq{between} our \trysq{dataset} and \trysq{the} 3D captioning dataset. Here we take one of the representative \trysq{3D captioning} works ScanRefer~\cite{chen2020scanrefer}. \trysq{Specifically}, we apply the Kolmogorov-Smirnov test on \trysq{the} \frysq{normalized} frequencies of nouns, verbs, and adjective tokens mentioned in the two datasets, following VQA~\cite{antol2015vqa}. \trysq{The results} prove that the underlying distributions of the two datasets differ significantly (p$<$1e-05). \trysq{In addition}, we perform the Paired T-test~\cite{ttest1997} and Anderson-Darling test~\cite{andersontest}, \trysq{which also show} a significant difference (p=1.4e-05 and p$<$1e-05, respectively).

All of these demonstrate that 3DQA captures information \trysq{that goes} beyond 3D captions by a wide margin and encourage us to collect our \textbf{fully human-annotated} dataset rather than relying on template-based generation from 3D captioning datasets.

\section{Method}
\label{sec:method}

\subsection{3DQA-TR Framework Overview}

\begin{figure*}[ht] 
\centering
\includegraphics[width=\textwidth]{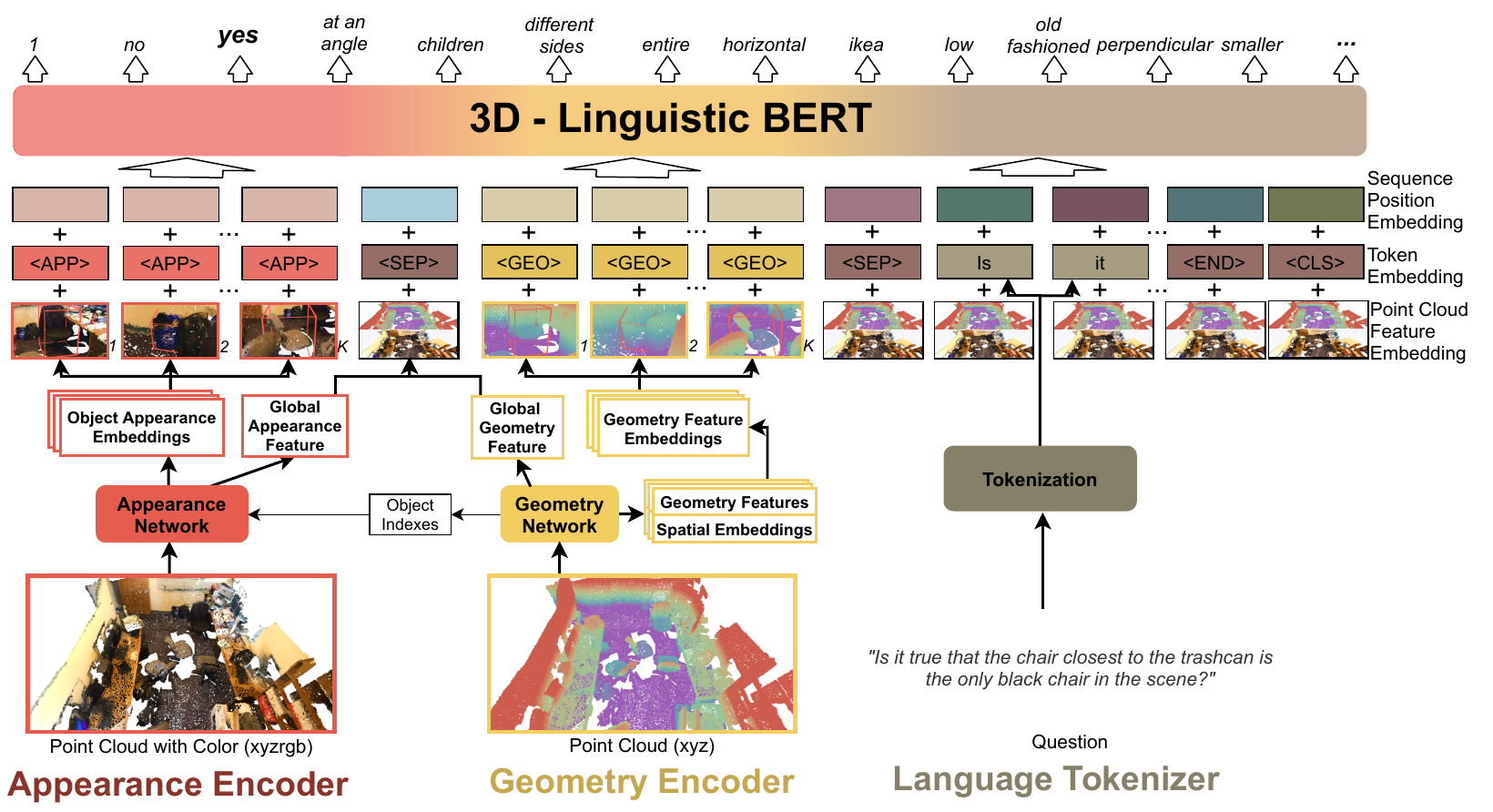}
\caption{Illustration of our pipeline. The 3DQA-TR framework has three jointly trainable parts: \trysq{the} appearance encoder (red), \trysq{the} geometry encoder (orange), and 3D-L BERT. The geometry encoder provides both structure information \trysq{about} each object and positional and scale information to model \trysq{the} spatial relationship between objects. The appearance encoder is used to extract appearance information. Given the encoded appearance and geometry together with question embedding, 3D-L BERT attends to the intra- and inter- modal interactions and predicts the answer.}
\label{pipeline}
\end{figure*}

\begin{figure*}[t] 
  \centering
  \includegraphics[width=0.9\textwidth]{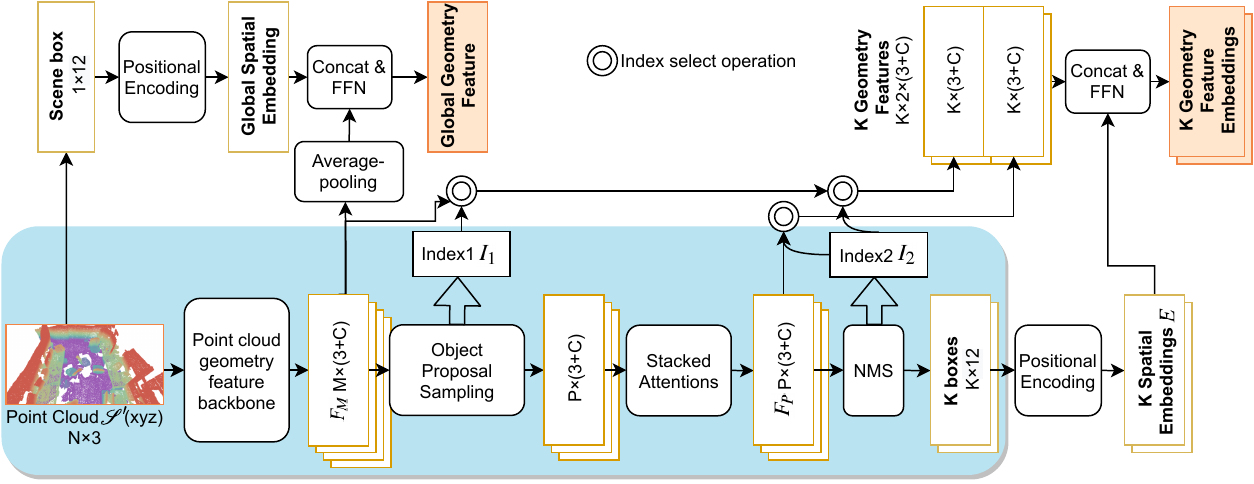}
  \caption{Illustration of the extraction of geometry features, spatial embeddings of \trysq{the} detected objects, and \trysq{a} global geometry feature by our geometry encoder.}
  \label{geobranch}
\end{figure*}

The overall framework of 3DQA-TR is summarized in Fig.~\ref{pipeline}. Given an input 3D scene and a corresponding question, our 3DQA-TR regresses the answer as a classification problem over all candidate answers. The candidate answers are collected from \textit{training} split, considering both subject agreement and answer confidence.

In the geometry encoder, given a scene point cloud $\mathcal{S}' \in \mathbb{R}^{N\times3}$ (xyz without \trysq{RGB}), a geometry network detects $K$ 3D object proposals and extracts the geometry feature of each object and spatial embedding for relationship modeling. %
In the appearance encoder, given the point cloud $\mathcal{S} \in \mathbb{R}^{N\times3}$ (xyz with \trysq{RGB}), an appearance network extracts color features for each detected object to support appearance-related questions. 
The extracted geometry and appearance information, as well as question embedding are further embedded as appearance, geometry and linguistic elements. Taking these elements as input, 3D-L BERT conducts both intra- and inter-modal interactions and outputs the predicted answer. Following the standard practice, our model is driven by \srysq{cross-entropy} loss.

\subsection{Geometry Encoder}

We first introduce the extraction of geometry features and spatial embedding in the geometry network, as shown in \Fref{geobranch}. %
 Here, \textit{\trysq{g}eometry features} are defined as features of each object extracted from one 3D object detection network. The pipeline of the detection network is shown in the blue region. In detail, following the common practice~\cite{Shi_2019_CVPR,qi2019deep,liu2021}, we take point clouds of the entire \srysq{scene} %
$\mathcal{S}' \in \mathbb{R}^{N\times3}$ (xyz without \trysq{RGB}) as input and feed it into a point cloud geometry feature backbone network (PointNet++~\cite{qi2017pointnet++} in our implementation), which extracts the feature representations $F_M$ of a subset of $M$ points with features of $3+C$-dim. Then $P$ object candidates are generated by initial object proposal subsampling, whose sampling index is denoted by $I_{1}$. The refined representations $F_P$ of $P$ points will be extracted by the stacked attentions if the detector backbone is Group-Free~\cite{liu2021}. 
The final 3D bounding boxes of \trysq{a} maximum \trysq{of} $K$ detected objects are generated by 3D NMS, whose sampling index is denoted by $I_{2}$. In most seniors, the object number is less than $K$, and we will apply zero padding.
To extract the geometry features of each object, the object \trysq{indices} of the above two sampling steps are \trysq{used} to trace back to the corresponding feature representations in the backbone model and the refined representation in the detector for each object. In detail, we \trysq{obtain the} object feature representations by $Idx(Idx(F_M,I_{1}),I_{2})$ and refined object representations by $Idx(F_P,I_{2})$, where $Idx(F,I)$ denotes index select operation on feature $F$ with index $I$. These are served as geometry features. 

\textit{Spatial embedding} is to explicitly incorporate inter-object relationship to handle a number of questions about spatial relationship in 3DQA. To this end, we utilize point locations, scale, and bound information of each detected object. In \srysq{detail}, we characterize each predicted bounding boxes into a $12$-d\trysq{imensional} vector. For box $k \in [1,K]$, the vector is formed as 
\begin{align*}
v^k = ( & \frac{x_c^k}{X},\frac{y_c^k}{Y},\frac{z_c^k}{Z},\frac{dx^k}{X},\frac{dy^k}{Y},\frac{dz^k}{Z}, 
\\
& \frac{x_{min}^k}{X},\frac{x_{max}^k}{X},\frac{y_{min}^k}{Y},\frac{y_{max}^k}{Y},\frac{z_{min}^k}{Z},\frac{z_{max}^k}{Z}) \in \mathbb{R}^{12}    \srysq{,}
\end{align*}
\srysq{where} $x_c^k,y_c^k,z_c^k$ are \trysq{the coordinates} of the center of th\trysq{e} box, and $dx^k,dy^k,dz^k$ are \trysq{the} scales of this box, and $x_{min}^k,x_{max}^k$ are \trysq{the} minimum and maximum $x$ values of this box. To ensure the scale invariance, they are divided by the scales of this scan in $X,Y,Z$ directions. \trysq{Then,} positional encoding~\cite{vaswani2017attention} is applied to the vector.
\begin{equation}
\left\{\begin{array}{l}
\textrm{PE}(v^k, 2 i)=\sin \left(v^k / 1000^{2 i / d_{\textrm{model }}}\right) \srysq{,}\\
\textrm{PE}(v^k, 2 i+1)=\cos \left(v^k / 1000^{2 i / d_{\textrm{model }}}\right) \srysq{,}
\end{array}\right.
\end{equation}
where $i\in[0,...,d_{\text {model }}/2)$ and $d_{\text {model }}$ is the target embedding dimension.
\trysq{Thus,} for each object, we \trysq{obtain} a high-dimensional representation $e^k = PE(v^k) \in \mathbb{R}^{12\times d_{\text {model }}}$ for the location, scale and bounds, which is \trysq{considered} as the spatial embedding. \ysq{The spatial embeddings of $K$ objects is formed by the set of the embeddings of all objects, $E = \{e^k\}^K_{k=1}$. }

Finally, we obtain geometry features and spatial embeddings, both of which represent \trysq{the} geometry information of $K$ objects. Before feeding them into 3D-L BERT, they will form geometry feature embeddings of $K$ objects, by concatenation and feeding into a feed-forward network.

To extract \trysq{a} global geometry feature, we first apply positional encoding to the entire scene's box vector to obtain a global spatial embedding $v=(x_c,y_c,z_c,dx,dy,dz,x_{min},x_{max},y_{min},y_{max},z_{min},z_{max}) \in \mathbb{R}^{12}$. \trysq{This} is \trysq{then} concatenated with the global feature, which is extracted by average-pooling on the feature representations $F_M$. After applying a feed-forward network, the global geometry feature is obtained.

\subsection{Appearance Encoder} 

To answer appearance-related questions, we \trysq{must} extract color features for each object proposal. However, 3D object detectors tend to ignore the RGB information and focus \trysq{primarily on} the geometry information. To mitigate this issue, we design a separate appearance encoder to capture \trysq{the} color information of \trysq{the} object proposals and pre-train this appearance encoder on color-related questions. 

Given a color point cloud $\mathcal{S} \in \mathbb{R}^{N\times6}$, the appearance network outputs \srysq{the} appearance features of each object proposal. In detail, we first use a PointNet++ network ~\cite{qi2017pointnet++} to extract features $F_\mathcal{S}$ of all points. \trysq{Because} objects are detected in the geometry encoder via the initial object proposal subsampling of index $I_{1}$ and 3D NMS of index $I_{2}$ as described in the previous section, here we can use these \trysq{indices} to select the appearance features of each detected object. It is denoted by $F_{app} = Idx(Idx(F_\mathcal{S},I_{1}),I_{2})$ and will form appearance features $F_{app}$ of $K$ objects. To extract global appearance features, we apply average-pooling on all point features.

To enforce the appearance extraction, we pre-train the appearance network on a synthetic question answering dataset tailored for appearance information extraction. 

This dataset is oriented for appearance information extraction -- the question and answer pairs only \trysq{concern the} colors of the objects. The question generation \trysq{templates} are \trysq{of} only two types `What color is the $<$ObjName$>$?' (single object) or `What color are the $<$ObjName$>$?' (multi\trysq{ple} objects), \trysq{to} avoid overfitting on \trysq{the} language prior. In detail, to generate the questions, we \trysq{use} the annotated objects in ScanNet~\cite{dai2017scannet}, \trysq{except} the wall, floor, and ceiling classes. To generate the answers, we first collect all points of the corresponding object\trysq{s} with \trysq{the} instance masks provided by ScanNet. \trysq{T}hen, we find \trysq{the color name of each point} by selecting \trysq{the} closest color \trysq{from} the \trysq{17} named CSS 2.1 colors. Then, using all the color names for each point in this object, we vote for one to two colors. For a single object, \trysq{this color} is the final answer. For multi\trysq{ple} objects, we combine the set of all the color names of \trysq{the} individual objects to form the final answer. The ablation experiments in the following section will demonstrate the effectiveness of the appearance encoder and the pretraining. We note that there is no overlap between the synthetic dataset and the ScanQA. The number of questions in the synthetic dataset is 3710, 982, 0 for train, validation and test splits\srysq{, respectively}.

\subsection{3D-Linguistic BERT}

\trysq{The} BERT~\cite{devlin2018bert} model has prove\trysq{n} its \trysq{ability} to aggregate and align multi-modal information in visual-language tasks.  Inspired by this, we \trysq{propose} 3D-L BERT to aggregate the multi-modal information from \trysq{the} appearance encoder, geometry encoder, and linguistic tokenizer, to make the final answer prediction.
The 3D-L BERT \trysq{input} is mainly composed of three types of elements: appearance, geometry, and \srysq{linguistics}. \trysq{S}ome auxiliary elements for modality separation or information summarization (classification token) are also added. As shown in Fig.~\ref{pipeline}, each element is the sum of Point Cloud Feature Embedding, Token Embedding, and Sequence Position Embedding. 

\textit{Point cloud feature embedding} is the geometry embedding from a point cloud or appearance embedding from a color point cloud. In \srysq{detail}, in geometry elements, they are the geometry feature embedding of $K$ objects. To gain the geometry feature embedding, we concatenate the geometry features with spatial embedding and feed them into a feed-forward network.
In appearance elements, they are \trysq{the} appearance embedding\trysq{s} of $K$ objects, which are formed by feeding the appearance features into a fully connected layer.
In linguistic and other auxiliary elements, they are the same embedding derived from a combination of both 
\trysq{the} appearance and geometry feature\trysq{s} of the scene. \trysq{They are} obtained by applying a linear layer after \trysq{concatenating the} global appearance and geometry features.

\textit{Token embedding} in the linguistic element is the WordPiece embedding of \trysq{the} question following the practice in BERT. For appearance and geometry elements, \trysq{special} $<$APP$>$ and $<$GEO$>$ token\trysq{s are} defined.
\textit{Sequence Position embedding} is to indicate the position of an element \trysq{among} all \trysq{the} elements. Specifically, \trysq{the} positions among the appearance elements are identical \trysq{because} they are not sequential, and so \trysq{are} geometry elements.

\begin{table*}[hbt!]
\setlength{\tabcolsep}{1.7mm}
\centering
\caption{Comparison of our method 3DQA-TR with the state of the art VQA methods LXMERT~\cite{tan2019lxmert}, VILBERT~\cite{lu2019vilbert} and 12-in-1~\cite{Lu_2020_CVPR} given question and the frame with the closest viewpoint and view direction to that of the ground-truth human answers. From left to right are overall performance, performances of different types of answers, and performances of different types of spatial-related questions. \ysq{The evaluation metric for the first 4 rows is exact match accuracy EM, and the metric for the last 4 rows is METEOR.}}
\scalebox{1.0}{
\begin{tabular}{@{}cccccccccc@{}}
\toprule
                            & \multicolumn{1}{c|}{All} & Number & Color & Y/N & \multicolumn{1}{c|}{Other} & aggregation & placement & spatial & viewpoint \\ \midrule
\textbf{EM} \\ 
Image+Q~1~\cite{tan2019lxmert}            & \multicolumn{1}{c}{31.15}    &  17.20   &  20.75      &  56.12     & \multicolumn{1}{c}{11.32} & 27.00 & 29.17 & 25.73 & 21.24 \\
Image+Q~2~\cite{lu2019vilbert}            & \multicolumn{1}{c}{31.26} & 20.43 & 18.87 & 55.58      & \multicolumn{1}{c}{11.58} & 28.50 & 30.56 & 25.15 & 20.35    \\
Image+Q~3~\cite{Lu_2020_CVPR}            & \multicolumn{1}{c}{32.37} & 21.51 & 20.75 & 56.91   & \multicolumn{1}{c}{12.36} & 27.50 & 31.94 & 26.32 & 23.00    \\
\textit{3DQA-TR}     & \multicolumn{1}{c}{\textbf{42.35}} & \textbf{40.86} & \textbf{35.85} & \textbf{64.63}     & \multicolumn{1}{c}{\textbf{21.58}} & \textbf{40.00} & \textbf{41.67} & \textbf{40.35} & \textbf{31.86}    \\
\midrule
\textbf{METEOR} \\ 
Image+Q~1~\cite{tan2019lxmert}            & \multicolumn{1}{c}{19.26} & 17.20 & 20.75 & 65.99   & \multicolumn{1}{c}{10.06} & 18.79 & 18.06 & 17.16 & 12.90     \\
Image+Q~2~\cite{lu2019vilbert}            & \multicolumn{1}{c}{18.88} & 20.43 & 18.87 & 62.06   & \multicolumn{1}{c}{9.09} & 17.85 & 17.39 & 16.79 & 11.69    \\
Image+Q~3~\cite{Lu_2020_CVPR}            & \multicolumn{1}{c}{19.24} & 21.51 & 20.75 & 62.55   & \multicolumn{1}{c}{9.20} & 17.38 & 18.52 & 17.13 & 11.91     \\
\textit{3DQA-TR}     & \multicolumn{1}{c}{\textbf{25.71}} & \textbf{40.86} & \textbf{35.65} & \textbf{73.49}    & \multicolumn{1}{c}{\textbf{15.87}} & \textbf{26.50} & \textbf{36.80} & \textbf{23.49} & \textbf{20.44}    \\
\bottomrule
\end{tabular}}
\label{tab:baseline}
\end{table*}

\begin{table*}[hbt!]
\setlength{\tabcolsep}{1.7mm}
\centering
\caption{\frysq{Additional comparison with various possible question answering methods.}
\srysq{The first 8 rows show related question answering methods divided into 4 groups based on their input settings: multi-view image, panoramic image, panoramic video, and spatial reasoning VQA. In each group, the second line is the performance of the state-of-the-art framework respectively, and the first line is of their backbone VQA model.
The 9-11 rows provide the performances of adapting the state-of-the-art Image+Q baselines for 3D+Q, by substituting their object detection backbone with 3D counterpart.}}
\scalebox{1.0}{
\begin{tabular}{@{}cccccccccc@{}}
\toprule
                            & \multicolumn{1}{c|}{All} & Number & Color & Y/N & \multicolumn{1}{c|}{Other} & aggregation & placement & spatial & viewpoint \\ \midrule
\rowcolor{LightGray}
BUTD~\cite{Anderson2017up-down}            & \multicolumn{1}{c}{ 27.27}    &  11.83    & 11.32        & 52.39       & \multicolumn{1}{c}{ 8.42} & 20.50  & 22.22  & 21.05  & 18.58  \\
\rowcolor{LightGray}
multi-view images~\cite{bansal2020visual}            & \multicolumn{1}{c}{ 28.82}    &  15.05    & 13.21        & 53.45       & \multicolumn{1}{c}{ 10.0} & 25.50  & 25.00  & 25.15  & 19.47  \\ \addlinespace
\rowcolor{LightLightGray}
MLB~\cite{kim2021mlb}            & \multicolumn{1}{c}{ 28.93}    &  15.05    & 13.21        & 52.93       & \multicolumn{1}{c}{ 10.26} & 26.50  & 23.61  & 23.39  & 18.58  \\
\rowcolor{LightLightGray}
panoramic image~\cite{9093452}            & \multicolumn{1}{c}{ 29.81} &   16.13& 18.87  &  52.93      & \multicolumn{1}{c}{ 11.84} & 27.50  & 25.00  & 24.56  & 18.58     \\ \addlinespace
\rowcolor{LightGray}
LXMERT~\cite{tan2019lxmert}            & \multicolumn{1}{c}{31.15}    &  17.20   &  20.75      &  56.12     & \multicolumn{1}{c}{11.32} & 27.00 & 29.17 & 25.73 & 21.24 \\
\rowcolor{LightGray}
spatial reasoning~\cite{banerjee2021weakly}             & \multicolumn{1}{c}{ 31.82} & 19.35  & 20.75  &  56.65   & \multicolumn{1}{c}{ 11.84} & 27.00  & 30.56  & 28.07  & 22.12     \\ \addlinespace
\rowcolor{LightLightGray}
BERT~\cite{devlin2018bert}           & \multicolumn{1}{c}{27.94} & 16.13 & 11.32 & 52.13   & \multicolumn{1}{c}{9.21} & 26.00 & 25.00 & 23.39 & 18.58    \\
\rowcolor{LightLightGray}
panoramic video~\cite{Yun2021PanoAVQA}           & \multicolumn{1}{c}{30.26} & 13.98 & 20.75 & 55.86   & \multicolumn{1}{c}{10.26} & 22.50 & 27.78 & 23.98 & 19.47    \\
\midrule
\cite{tan2019lxmert} with 3D detector          & \multicolumn{1}{c}{37.70} & 33.34 & 15.09 & 61.43   & \multicolumn{1}{c}{18.42} & 33.00 & 36.11 & 30.41 & 30.97    \\
\cite{lu2019vilbert} with 3D detector          & \multicolumn{1}{c}{38.69} & 35.48 & 16.98 & 62.23   & \multicolumn{1}{c}{19.21} & 34.50 & 36.11 & 31.58 & 28.32    \\
\cite{Lu_2020_CVPR} with 3D detector          & \multicolumn{1}{c}{37.03} & 32.26 & 13.21 & 61.70   & \multicolumn{1}{c}{17.11} & 31.50 & 34.72 & 30.99 & 29.20    \\
\midrule
\textit{3DQA-TR}     & \multicolumn{1}{c}{\textbf{42.35}} & \textbf{40.86} & \textbf{35.85} & \textbf{64.63}     & \multicolumn{1}{c}{\textbf{21.58}} & \textbf{40.00} & \textbf{41.67} & \textbf{40.35} & \textbf{31.86}    \\
\bottomrule
\end{tabular}}
\label{tab:relatedqabaseline}
\end{table*}

\begin{table*}[t]
\setlength{\tabcolsep}{1.8mm}
\centering
\caption{\ysq{EM of human answers with different input settings, indicating the high quality of our dataset and the importance of 3D scene comprehension.}}
\scalebox{1.0}{
\begin{tabular}{@{}c|ccccccccc@{}}
\toprule
 & \multicolumn{9}{c}{Human Accuracy EM (\%)}  \\
{Input}     & \multicolumn{1}{c|}{All} & Number & Color & Y/N & \multicolumn{1}{c|}{Other} & aggregation & placement & spatial & viewpoint \\ \midrule
{Question}     & \multicolumn{1}{c|}{\textcolor{black}{ 27.28  }}    &  \textcolor{black}{ 13.98}   &   \textcolor{black}{ 15.09}     &  \textcolor{black}{ 51.86}     &  \multicolumn{1}{c|}{\textcolor{black}{ 7.89}}      &   \textcolor{black}{ 16.00}                  &  \textcolor{black}{ 23.61}         &  \textcolor{black}{ 26.32}       &  \textcolor{black}{ 23.01}         \\
{Question+Image}     & \multicolumn{1}{c|}{\textcolor{black}{ 53.54}}    &  \textcolor{black}{ 47.31}   &  \textcolor{black}{ 49.06}      & \textcolor{black}{ 62.50}      & \multicolumn{1}{c|}{\textcolor{black}{ 46.84}}      & \textcolor{black}{ 46.00}                  &  \textcolor{black}{ 48.61}         & \textcolor{black}{ 49.12}        & \textcolor{black}{ 50.44}          \\
{Question+Scene} & \multicolumn{1}{c|}{\textcolor{black}{ 83.26}}    & \textcolor{black}{ 79.57}    &  \textcolor{black}{ 79.25}      &  \textcolor{black}{ 94.15}     & \multicolumn{1}{c|}{\textcolor{black}{ 73.95}}      &  \textcolor{black}{ 76.50}                  &  \textcolor{black}{81.94 }         & \textcolor{black}{82.46 }        & \textcolor{black}{82.30 }          \\\bottomrule

\end{tabular}}
\label{tab:accs}
\end{table*}

By adaptively integrating \textit{intra-modal} information inside each element type, 3D-L BERT enables the aggregation within geometry features or appearance features, and spatial relationship modeling among the objects represented by spatial embedding.
By aligning \textit{inter-modal} information among \trysq{the} three different element types, the linguistic information can selectively attend to \srysq{the} geometry or appearance clues of any object without including the redundant information. \trysq{In addition,} a scaling parameter for each embedding and each element is jointly learned to achieve a good balance.
The following ablation experiments \trysq{concerning} individual embedding will demonstrate the superiority of the design.

\section{Experiments}
\label{sec:experiments}

\begin{figure*}[hbt!] 
  \centering
  \includegraphics[width=0.92\textwidth]{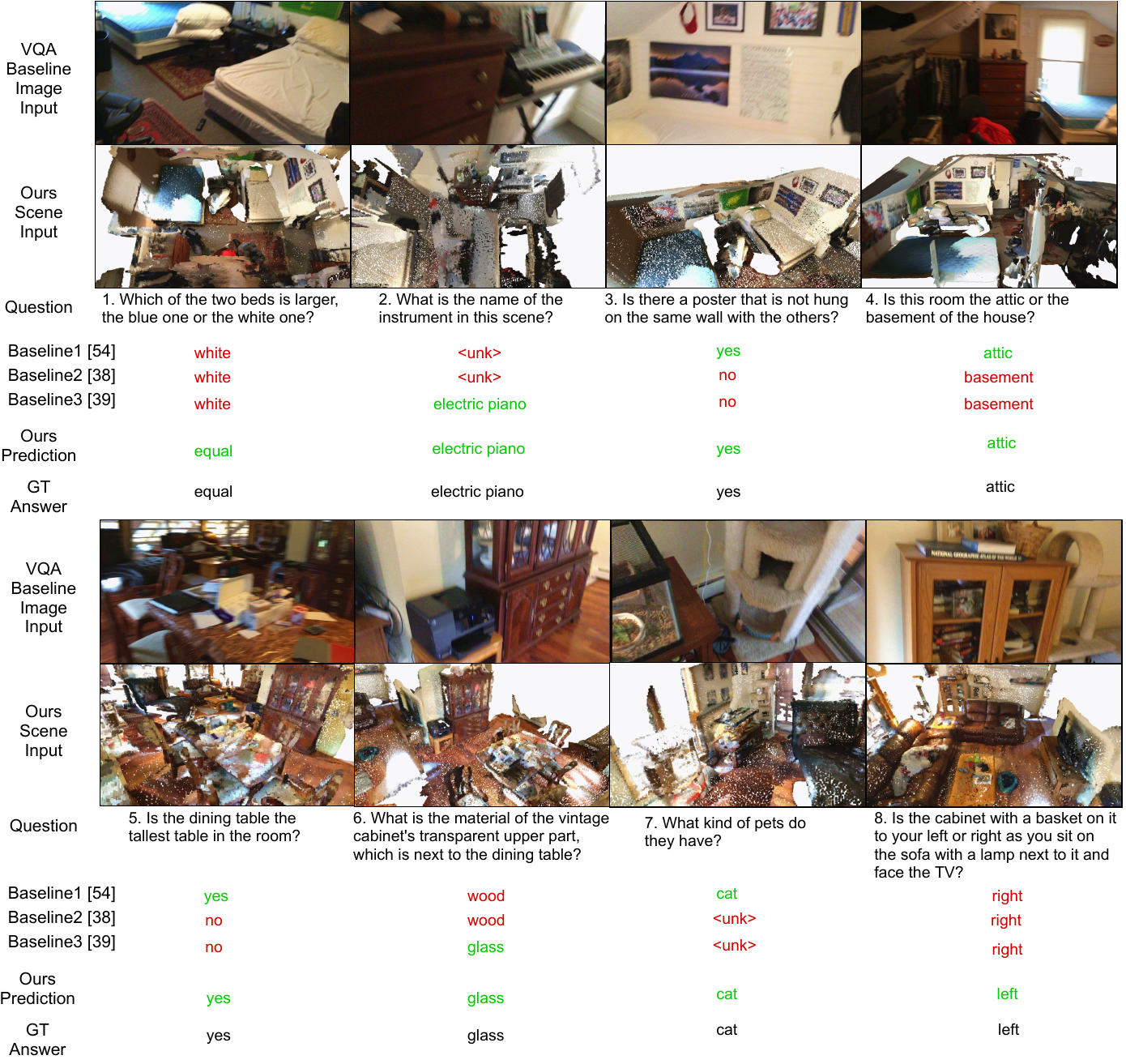}
  \caption{Example predictions from 2D VQA baselines and our 3DQA-TR, \trysq{compared to} human answers in ScanQA. The answers with $100\%$ EM scores are in green and \trysq{the others are} in red. \ysq{'$<$unk$>$' is the unknown tokens that represent the prediction is not in \trysq{the} candidate answers}, following the official implementations 
  ~\cite{tan2019lxmert,lu2019vilbert,Lu_2020_CVPR}.
  }
  \label{qualiresult}
\end{figure*}

We first describe the data preparation, evaluation metric and training. \trysq{Then}, we show both quantitative and qualitative results of our framework compared with state-of-the-art 2D VQA approaches, and \trysq{the} \trysq{performance} of human participants under different input settings. We also conduct ablation experiments to demonstrate the effectiveness of our major components.

\subsection{Data Preparation and Evaluation Metric} 
We split our data into train\trysq{ing}/val\trysq{idation}/test sets following the standard split~\cite{qi2019deep, liu2021} of ScanNet~\cite{dai2017scannet}, and ensure disjoint scenes \trysq{in} each split. %
\frysq{The number of questions is 8551, 609, 902 for train, validation and test splits\srysq{, respectively}.}
According to the type of the answers for each question, we split the test set into \trysq{four} sub-classes: Yes/No (Y/N), Color, Number, and \frysq{Others. }%
We use the top-1 accuracy, \trysq{called the} exact token match (EM), following the common practice in VQA~\cite{chou2020visual,docvqa_wacv,lobry2020rsvqa,biten2019scene,kafle2018dvqa,zhu2016visual7w,ren2015exploring,malinowski2014multi}, as the evaluation metric for the following experiments, \trysq{as} EM was introduced as a good evaluation metric in QA tasks, especially for single-word answers. We also include the METEOR~\cite{gurari2018vizwiz,chen2015microsoft,abadenkowski2014meteor} metric, which was designed \trysq{to evaluate} longer ph\trysq{r}ases, 
as more than $\frac{1}{4}$ of the answers in our ScanQA dataset exceed one word.

\subsection{Training}
The 3D-L BERT model is initialized by the official \frysq{PyTorch} pretrained B\trysq{ERT}-base model as described in~\cite{devlin2018bert}. 
Before joint\trysq{ly} fine-tuning \trysq{the} whole framework, the detector backbone (Group-Free~\cite{liu2021} by default) is pre-trained on the 3D object detection task\trysq{, a}nd the appearance network is pre-trained by question answering on the synthetic color-related questions. We train our framework on \trysq{the} train\trysq{ing} and val\trysq{idation} splits and report both quantitative and qualitative \trysq{results} on \trysq{the} test split.
We train \trysq{the model} on \trysq{four NVIDIA GeForce} 2080Ti GPUs with a total batch size of 64. For training, \trysq{the} AdamW optimizer is applied, with a base learning rate of 1e-8 for the feature extractor backbone PointNet++ and 5e-6 for the \trysq{remainder}, weight decay of 1e-4, learning rate warm\trysq{ing} up over the first 500 steps, and then follow\trysq{ing a} cyclical learning rate policy.

\subsection{Main Results}\label{sec:mainresults}

\noindent\textbf{Comparison with state-of-the-art VQA methods.} \trysq{Table}~\ref{tab:baseline} \trysq{compares} our method with the \frysq{state-of-the-art} 2D VQA methods LXMERT~\cite{tan2019lxmert}, VILBERT~\cite{lu2019vilbert} and 12-in-1~\cite{Lu_2020_CVPR}, which take images and questions as input. These Image+Q baselines are trained with questions \trysq{from} ScanQA and the corresponding video frames from ScanNet as input. To favor their performances, we chose the image whose viewpoint and view direction is closest to the viewpoint and view direction of the ground-truth human answer for \trysq{each} question, but our 3DQA-TR still outperforms \trysq{the} 2D VQA methods by a large margin in both EM and METEOR metrics, especially for spatial-related questions. %
This is expected because images are insufficient \trysq{for answering} 3DQA questions, even for human subjects.

\noindent\textbf{Human study.} To show \trysq{that} 3D scene comprehension is required for 3DQA, we also \trysq{conduct} a human-study experiment that examines human performance under three different input settings:
1) only the question\trysq{,}
2) the question and a single image that is closest to the viewpoint and view direction of the ground-truth human answer\trysq{, and}
3) the question and the corresponding 3D scene. 
For fairness, \trysq{the} questions are randomly chosen, and different settings of the same scene \trysq{are not} assigned to the same subject. In addition, participants were not given any prior knowledge of the dataset.
Table~\ref{tab:accs} shows that human participants who are given both \srysq{questions} and 3D scenes perform much better than \trysq{the} others, demonstrating the importance of 3D scene comprehension in 3DQA. We note that the human EM given both questions and 3D scenes is greater than $83\%$, indicating the high quality of our dataset.
Notably, the performance gap ($>20.0$\%) between Question+Image and Question+Scene is \trysq{larger} for spatial-related questions (\srysq{the} rightmost five columns). On the other hand, the performance gap between \trysq{the} human\trysq{s} and our 3DQA-TR also shows that there is still a \trysq{significant amount of} room to improve.

\noindent\textbf{Qualitative comparison with VQA methods.} \trysq{Fig}.~\ref{qualiresult} \trysq{shows} some qualitative examples of 3DQA-TR, \trysq{the} 2D VQA baseline, and the ground-truth human answers in our test set, for questions of different types -- questions about ``spatial'' concepts (1), ``placements'' (6, 8), ``viewpoint and navigation'' (8), and ``aggregation'' (2, 5). Commonsense reasoning ability is also required \trysq{to answer} \trysq{Q}uestions (2, 4, 6, 7) in our dataset, which further verifies the variance of our dataset and the robustness of our method to different question types. In all the questions of all types, our predictions are scored 100\% correct, \trysq{demonstrating} the \trysq{capability} of the proposed framework for both appearance and geometry comprehension\frysq{. }Specifically, \trysq{Q}uestion 8 shows popular scenarios where VQA methods fail for specifying viewpoint or doing navigation. \trysq{In addition, Q}uestion 1, which requires a comparison of spatial information, \trysq{namely}, \trysq{the} sizes of two beds, is a common \srysq{scenario} in ScanQA. However, we can see all the state-of-the-art 2D VQA baselines end up with wrong predictions. \trysq{This result} is expectable because of the perspective projecting nature of 2D images, which renders the size of the white one larger than the blue one. In contrast, the prediction of our model is correct. We attribute it to the superiority of our geometry encoder \trysq{over the 2D-based baselines, as our encoder} explicitly extracts and incorporates geometry embeddings of the objects’ spatial information in 3D \srysq{scans}.

\begin{table*}[hbt!]
  \setlength{\tabcolsep}{1.8mm}
  \centering
    \caption{Ablations of our method 3DQA-TR with components \trysq{of} deduction, \trysq{spatial embedding settings}, element design, appearance encoder pre-\trysq{training} and different detection backbones. The evaluation metric for the first \trysq{eight} rows is EM, and the metric for the last \trysq{eight} rows is METEOR.}
  \scalebox{1.0}{
  \begin{tabular}{@{}cccccccccc@{}}
  \toprule
                                & \multicolumn{1}{c|}{All} & Number & Color & Y/N & \multicolumn{1}{c|}{Other} & aggregation &  placement & spatial & viewpoint \\ \midrule
  \textbf{EM} \\ 
  Qonly            & \multicolumn{1}{c}{27.94} & 16.13 & 11.32 & 52.13   & \multicolumn{1}{c}{9.21} & 26.00 & 25.00 & 23.39 & 18.58    \\
  Geo+Q     & \multicolumn{1}{c}{40.58} & 37.63 & 13.20 & 64.89    &  \multicolumn{1}{c}{21.05} & 37.50 & 40.28 & 38.01 & 33.63   \\
  App+Q     & \multicolumn{1}{c}{33.59} & 29.03 & 30.19 & 54.79    & \multicolumn{1}{c}{14.21} & 33.50 & 25.00 & 26.32 & 21.24     \\ 
  NoSpaEmbedding     & \multicolumn{1}{c}{38.02} & 33.33 & 32.08 & 60.64    & \multicolumn{1}{c}{17.63} & 36.00 & 31.94 & 28.65 & 24.78     \\ 
  OneElementForAll     & \multicolumn{1}{c}{38.91} & 35.48 & 33.96 & 60.10    & \multicolumn{1}{c}{19.47} & 35.00 & 41.67 & 35.67 & 30.09    \\ 
  AppFromScratch     & \multicolumn{1}{c}{40.80} & 37.63 & 15.09 & 64.63  & \multicolumn{1}{c}{21.58} & 38.50 & 38.89 & 39.77 & 30.97   \\ \addlinespace
  \textit{\textit{3DQA-TR (backbone B)}}    & \multicolumn{1}{c}{41.79} & 41.94 & 39.62 & 63.30   & \multicolumn{1}{c}{20.79} & 39.00 & 40.28 & 39.18 & 30.97      \\
  \textit{\textit{3DQA-TR}}     & \multicolumn{1}{c}{42.35} & 40.86 & 35.85 & 64.63   & \multicolumn{1}{c}{21.58} & 40.00 & 41.67 & 40.35 & 31.86      \\
  \midrule
  \textbf{METEOR} \\ 
  Qonly            & \multicolumn{1}{c}{17.54} & 16.13 & 11.32 & 63.01    & \multicolumn{1}{c}{8.83} & 18.43 & 16.16 & 15.19 & 11.44    \\
  Geo+Q     & \multicolumn{1}{c}{24.45} & 37.63 & 13.21 & 72.56   &  \multicolumn{1}{c}{15.60} & 25.16 & 36.44 & 22.02 & 20.94    \\
  App+Q     & \multicolumn{1}{c}{20.60} & 29.03 & 30.02 & 61.99     & \multicolumn{1}{c}{11.60} & 22.71 & 22.08 & 15.56 & 13.97    \\ 
  NoSpaEmbedding     & \multicolumn{1}{c}{23.35} & 33.33 & 31.89 & 66.81      & \multicolumn{1}{c}{13.74} & 24.14 & 27.78 & 17.53 & 15.79    \\ 
  OneElementForAll     & \multicolumn{1}{c}{23.72} & 35.48 & 33.77 & 68.12     & \multicolumn{1}{c}{14.66} & 23.55 & 36.27 & 20.80 & 19.57   \\ 
  AppFromScratch     & \multicolumn{1}{c}{24.52} & 37.63 & 15.09 & 71.93    & \multicolumn{1}{c}{15.60} & 25.64 & 35.19 & 23.68 & 19.37    \\ \addlinespace
  \textit{\textit{3DQA-TR (backbone B)}}    & \multicolumn{1}{c}{25.35} & 41.94 & 39.40 & 71.65    & \multicolumn{1}{c}{15.03} & 25.83 & 35.86 & 22.84 & 19.70       \\
  \textit{\textit{3DQA-TR}}     & \multicolumn{1}{c}{25.71} & 40.86 & 35.65 & 73.49     & \multicolumn{1}{c}{15.87} & 26.50 & 36.80 & 23.49 & 20.44     \\ 
  \bottomrule
  \end{tabular}}
  \label{tab:ablation}
\end{table*}

\noindent\textbf{\frysq{Comparison with related question answering methods.}} 
To further illustrate the advantage of question answering with \trysq{the} point cloud and the effectiveness of our proposed method, in the first \trysq{eight} rows of ~\Tref{tab:relatedqabaseline}, we compare \trysq{the results of four} groups of related question answering methods with inputs of multi-view image\trysq{s}, panoramic image\trysq{s}, panoramic video, and spatial reasoning VQA. In each group, the second lines are the performance of the state-of-the-art frameworks with each input setting, respectively, and the first lines are \trysq{their} backbone VQA models. The metric \trysq{is EM}.
\trysq{Our} 3DQA-TR outperforms related question answering works by a large margin, especially in spatial \trysq{questions} (\trysq{an improvement of} more than 12.28\%) and aggregation (\trysq{an improvement of} more than 12.5\%). Although \trysq{the other methods had} performance gains compared with their backbones, they still fail to solve this 3DQA \srysq{task} as \srysq{effectively} as \srysq{our} framework. These results confirmed the effectiveness of the proposed framework, which we attribute to \trysq{its} 3D perception ability and the advantage of the point cloud, which directly captures more direct access to rich\trysq{,} accurate 3D spatial information.

\trysq{Rows} 9 to 11 of ~\Tref{tab:relatedqabaseline} \trysq{show the results} of adapting \trysq{the} Image+Q baselines for 3D+Q, by substituting \trysq{the} object detection backbone with \trysq{a} 3D counterpart. Although \trysq{the state-of-the-art methods'} performance with the 3D detector is better than their performance without the 3D detector in the paper, they still fail to outperform our method, demonstrating that these methods are incapable of solving \trysq{3DQA}. These results further demonstrate the effectiveness of the proposed framework, and we owe it to the superiority of the design of the framework with three different element types: appearance, geometry and language elements\trysq{. These elements} are specifically created for solving 3DQA task, especially with the extraction of appearance features from the pre-trained appearance encoder, the extraction of geometry features and spatial embeddings, and the inter- intra- fusion among \trysq{the} three \srysq{modalities}.

\subsection{Ablation Experiments}
\label{exp:ablation}

\subsubsection{Component Validation} 
In the first three rows of Table~\ref{tab:ablation}, we show \trysq{the} ablation with\trysq{ components removed from} our framework. 
In the ``Qonly" row, both the appearance and geometry encoders, as well as the appearance and geometry element\trysq{s} \trysq{of} 3D-L BERT are removed. Thus \trysq{the model} works like the original BERT model and provides a lower bound for solving 3DQA with only questions as input. It has the worst overall performance of all the baselines, especially \trysq{considering} the drop compared with the full framework (from $42.35\%$ to $27.94\%$). This \frysq{indicates} that information beyond \trysq{the} question is required, and \trysq{it} demonstrates the reasoning ability of our framework \trysq{rather than} relying on language bias. We also noticed an interesting phenomenon: the overall performance is slightly \trysq{better} than that of \trysq{the} humans in the ~\ref{tab:accs}. 
It is \trysq{unsurprising} because the network can learn the prior knowledge of the data distribution while the tested people are not the annotators of our dataset and have no such prior.

\trysq{On} the ``Geo+Q" row, the appearance element is removed from the input. We can \trysq{see} an overall performance decline compared to the full framework, especially in ``\trysq{c}olor" split (from $35.85\%$ to $13.20\%$), which further demonstrates the \trysq{need for} appearance information for 3D question answering. \trysq{However, the performance is clearly better than with} the ``Qonly" setting, especially in all the spatial-related questions, \trysq{such as} ``spatial'' \srysq{(}$23.39\%$ to $38.01\%$), ``placement'' \srysq{(}$25.00\%$ to $40.28\%$), ``viewpoint and do navigation'' \srysq{(}$18.58\%$ to $33.63\%$), 
and ``aggregation'' \srysq{(}$26.00\%$ to $37.50\%$) subset\trysq{s}, which we owe to the superiority of the geometry encoder.  %

\trysq{On} the ``App+Q" row, we removed \trysq{the} geometry element from the input. The performance, especially for the spatial-related questions, \trysq{such as the} ``spatial''
and ``placement'' subset\trysq{s}, \trysq{is worse than} \srysq{the} full model, indicating that the geometry information really matters. \trysq{However, the performance is better than the} ``Qonly" setting, \trysq{and the} performance gain in the ``color'' \trysq{split} ($11.32\%$ to $30.19\%$) is also noticeable, demonstrating the capability of the appearance encoder.%

\subsubsection{Appearance Questions Pre-training} 
In this ablation study, we demonstrate the design of the proposed pretraining of \trysq{the} appearance encoder in Table~\ref{tab:ablation}. The performance with \trysq{the} appearance encoder trained from scratch is shown in the ``AppFromScratch" row. 
In contrast, for our default design, the appearance encoder is pretrained on a generated question-answering dataset. On this generated dataset, our \trysq{model's} accuracy is $45.03\%$, which is much higher than $25.74\%$, the lower bound \trysq{of predicting} the most frequent answer to \trysq{the} question (e.g., ``What color is the wall?''-``White''). After applying the pretraining weight and then joint\trysq{ly}-training on the ScanQA dataset (``3DQA-TR'' row), the performances \trysq{on} all splits are better than \trysq{on} the \trysq{counterpart} trained from scratch (\trysq{EM improved from $40.80\%$ to $42.35\%$), especially in \trysq{the} ``color'' split (EM improved from $15.09\%$ to $35.85\%$).}

\subsubsection{Element Design} 
We show \trysq{the} experimental results to demonstrate that embedding \trysq{the} geometry and appearance into one element will cause \trysq{a} redundancy problem. In the ``OneElementForAll" row of Table~\ref{tab:ablation}, we embed\trysq{ded the} geometry feature, spatial embedding, and appearance feature into one element by \trysq{concatenation, rather than} embedding them into two separate elements. We can see a clear performance decline in the ``OneElementForAll" setting for all splits, for both ``Color'' questions (EM from $35.85\%$ to $33.96\%$) and spatial-related questions (e.g., EM of ``spatial'' from $40.35\%$ to $35.67\%$). This \frysq{demonstrates} the \trysq{need to} separat\trysq{e} geometry and appearance elements \trysq{in our design}.

\subsubsection{Spatial Embedding} 
Here we conduct one ablation on the geometry encoder to remove the spatial embedding\trysq{, leaving} only the geometry \trysq{features}. The results are shown in the ``NoSpaEmbedding" row of Table~\ref{tab:ablation}. The overall \trysq{performance} of both metrics are worse than that of the full framework (EM from $42.35\%$ to $38.02\%$, METEOR from $25.71\%$ to $23.35\%$). \trysq{Specifically, there is a significant} decline \trysq{in the} EM \trysq{for} spatial-related questions -- ``spatial'' ($40.35\%$ to $28.65\%$), ``placement'' ($41.67\%$ to $31.94\%$), \trysq{and} ``viewpoint'' ($31.86\%$ to $24.78\%$)
. \trysq{In addition}, the drop \srysq{in} \trysq{the} METEOR performance in these questions can \trysq{clearly} be \trysq{seen}.
The performance gaps in \trysq{the} spatial-related questions \frysq{indicate} the importance of modeling \trysq{the} spatial information in bounding boxes and \trysq{the} \srysq{relationships} among them via spatial embeddings.

\subsubsection{Different Detector} 
To demonstrate the generalizability of our 3DQA-TR framework, we replace the default detector Group-Free~\cite{liu2021} with another backbone network Vote~\cite{qi2019deep}, denoted as 3DQA-TR (backbone B). 
As shown in Table~\ref{tab:ablation}, the overall performances and \trysq{the performance} in each \trysq{subclass} of 3DQA-TR (backbone B) \srysq{are} comparable with 3DQA-TR with the default backbone Group-Free \trysq{using} both metrics, \trysq{demonstrating} the generalizability of the proposed framework to different detectors.

\section{Conclusion}
\frysq{This paper extends the 2D VQA task into its 3D counterpart, the 3DQA task. }
\trysq{To} answer questions about a real-world 3D scene, 3DQA \trysq{must} understand both appearance and 3D geometry. To this end, we propose a novel end-to-end 3DQA framework 3DQA-TR by designing a multi-modal 3D-Linguistic BERT. The appearance and geometry information are encoded by two separate encoders, and then fed into the BERT together with the question embedding to predict the final answers. To support the 3DQA task, we also develop a new dataset, ``ScanQA" for this task, which contains \ysq{$10,062$ questions and answers} from $806$ scenes. Extensive experiments and analyses have demonstrated the superiority of our 3DQA-TR over existing VQA baselines.

\section{Limitations} \label{sec:limit}
Though we have tried our best to increase the diversity of the newly built ScanQA dataset, it is still not as diverse as existing large-scale image VQA datasets. This is mainly because ScanQA is built upon the ScanNet dataset, which only contains 806 indoor scenes. \trysq{The l}imited diversity of the dataset will \trysq{impair} the generaliza\trysq{bility} and robustness of the model. For example, our model tends to give \trysq{incorrect} answers to \trysq{questions} about humans, as humans are \trysq{scarce} in this dataset. To \trysq{address} this problem, we are currently investigating large-scale 3DQA data generation and developing a multi-modality data augmentation technique for 3DQA. Another possible solution is to exploit extra large-scale 2D VQA data by applying knowledge distillation techniques.
\trysq{In addition}, as discussed in \Sref{sec:mainresults}, in our ScanQA dataset, \trysq{common sense} regarding entity names and reasoning ability is necessary, but it is still unclear how far the model can go beyond perception and recognition.

  \section*{Acknowledgments}
    We would like to thank the reviewers for their constructive feedback. This work was supported by the Hong Kong Research Grants Council (RGC) GRF Scheme under grant CityU 11216122.

\ifCLASSOPTIONcaptionsoff
  \newpage
\fi

\newpage
In this supplemental material, we introduce the data collection process and the web-based user interface, as well as perform some additional analyses on our dataset. Then we show the details of the comparison with VQA provided in our main paper. In the next section, we provide more implementation details. Finally, we discuss more about the limitation of our work to further inspire future work.

\begin{figure*}[hbt!] \centering
    \includegraphics[width=0.49\textwidth]{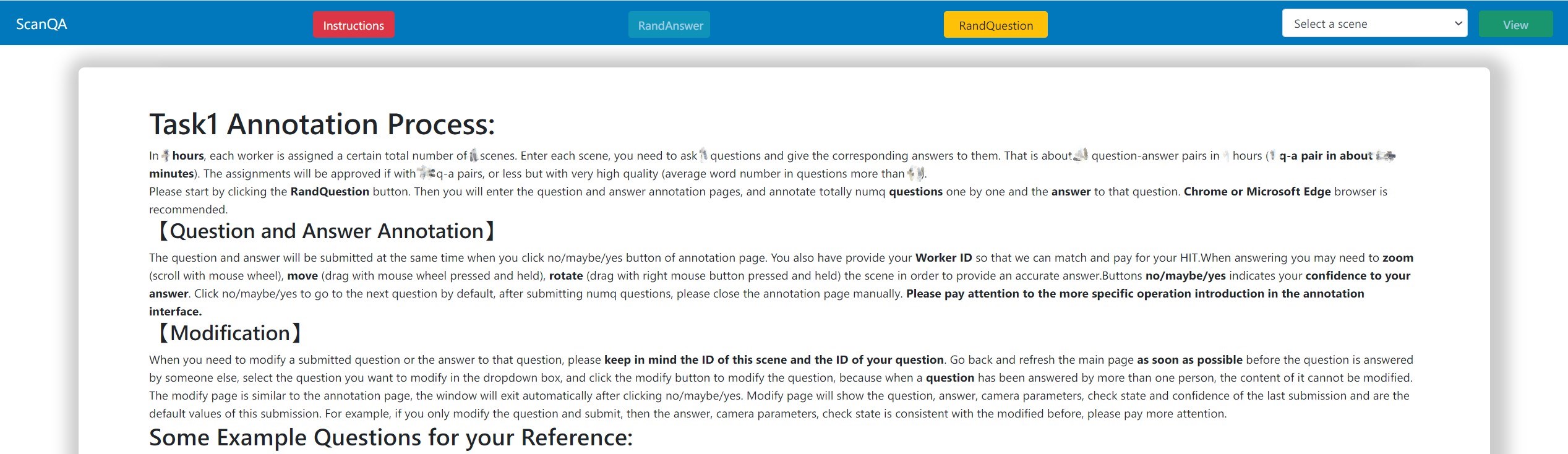}
    \includegraphics[width=0.49\textwidth]{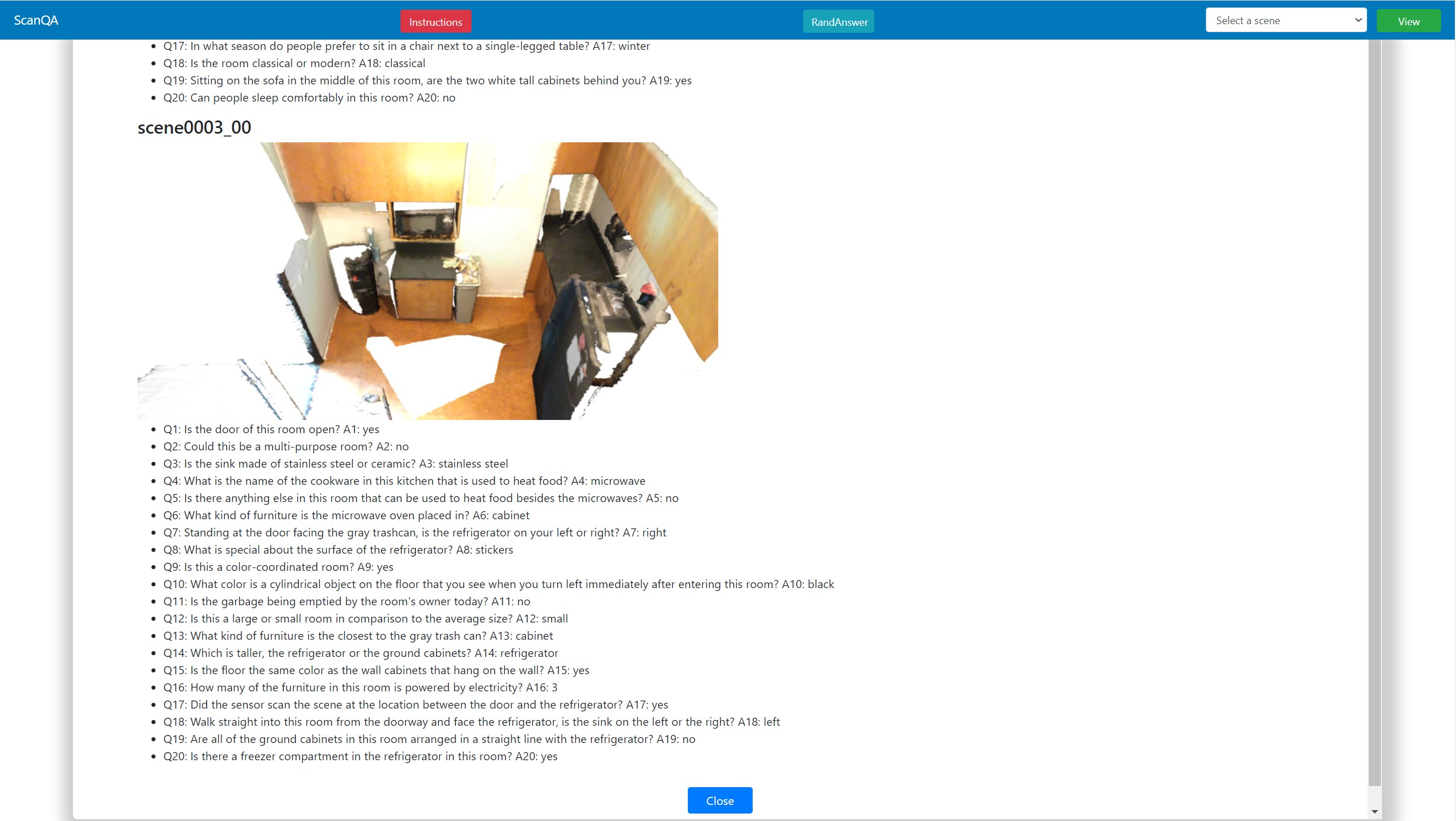}
    \caption{Our instruction page for the Amazon Mechanical Turk (AMT) interface, which shows the annotation pipeline and interesting questions as prompts.} \label{supp-webmain}
\end{figure*}

The detailed instruction page and annotation page for collecting questions and answers is Illustrated in Fig. 3 in our main paper.

\begin{figure*}[hbt!] \centering
  \includegraphics[width=0.49\textwidth]{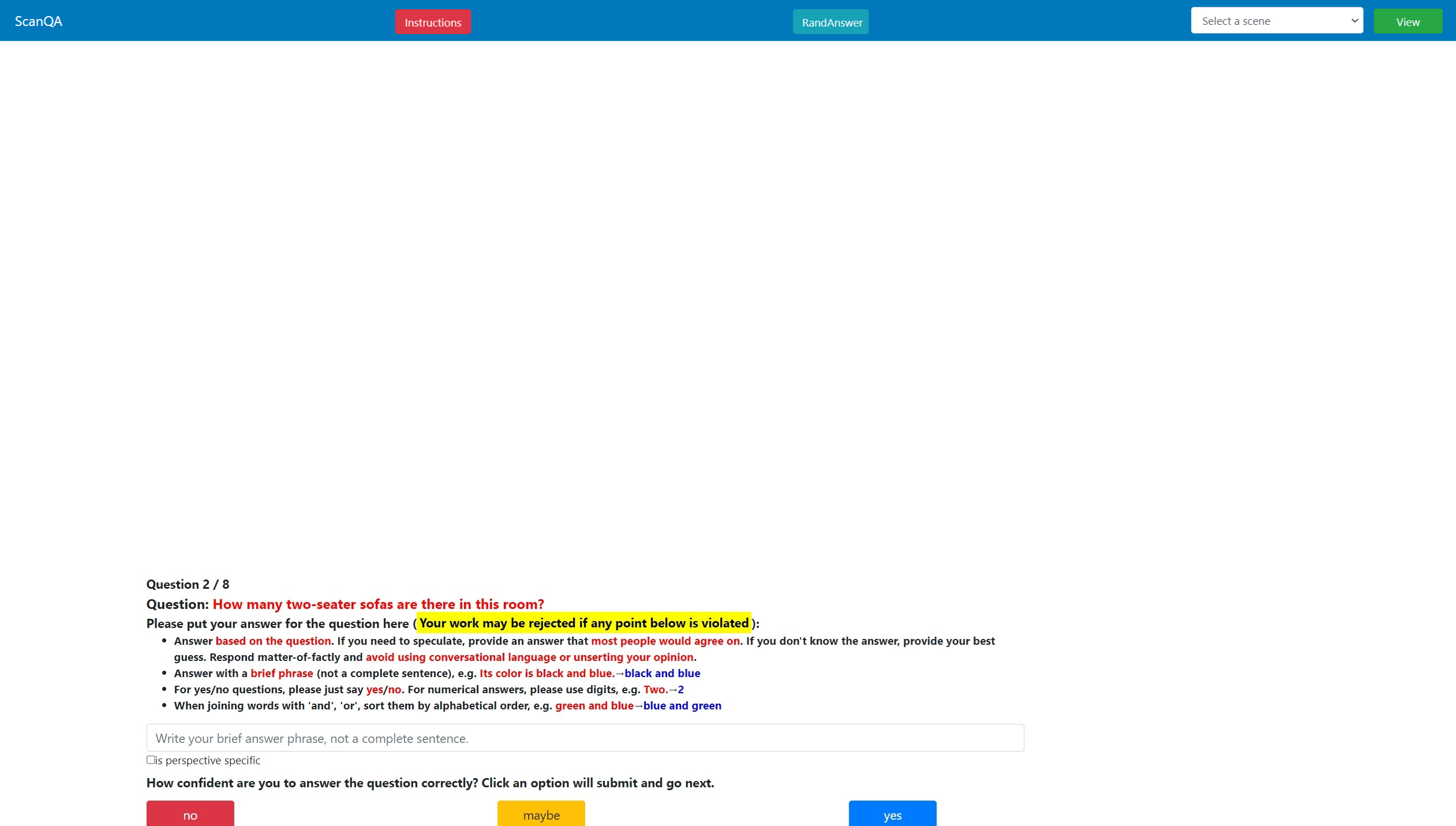}
  \includegraphics[width=0.49\textwidth]{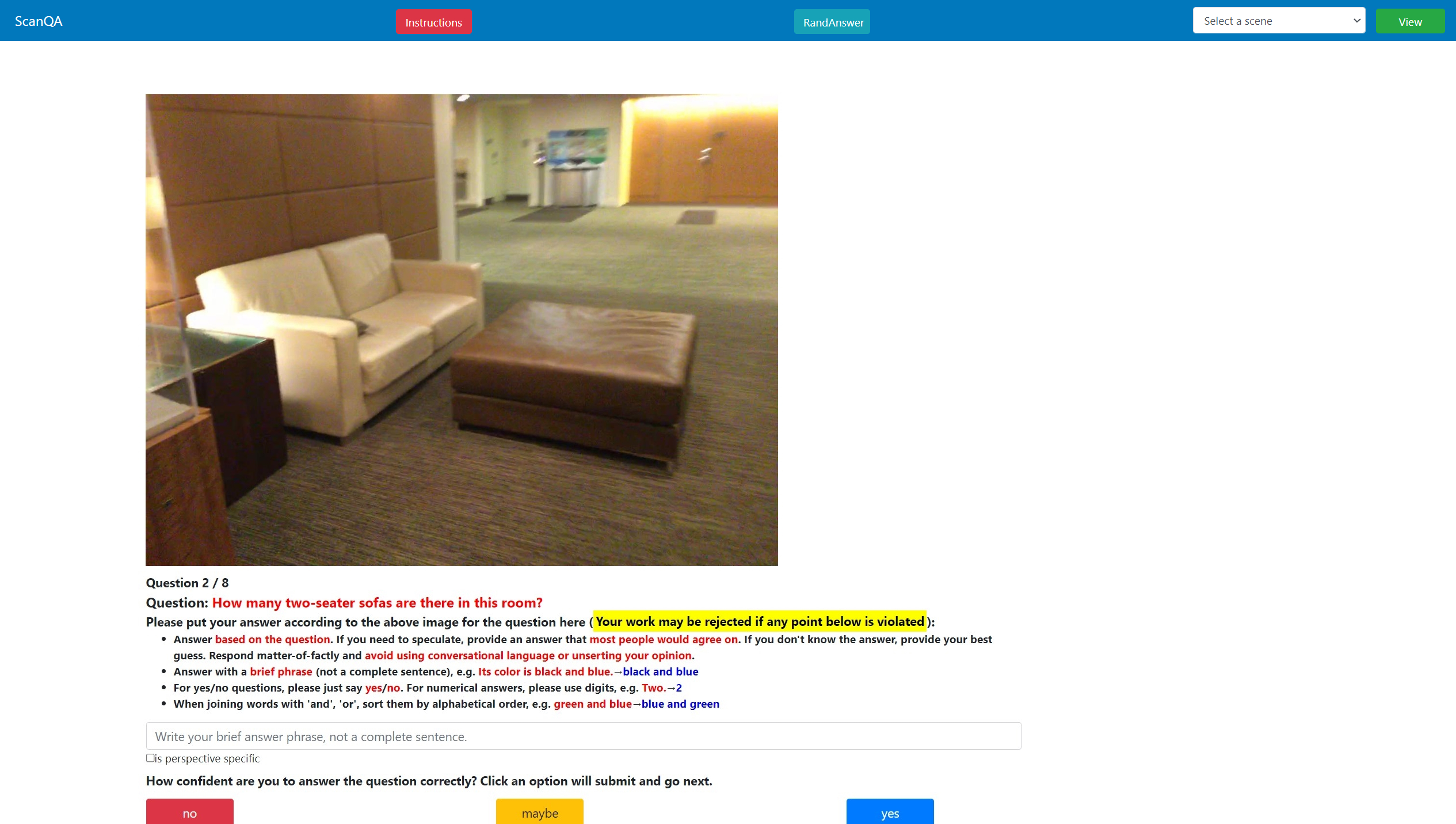}
  \caption{Our AMT interface for collecting the answer for input setting `Question' and `Question+Image' in our main paper.} \label{supp-webQonlyandQimg}
\end{figure*}

\begin{figure*}[hbt!] 
\centering
\includegraphics[width=0.49\textwidth]{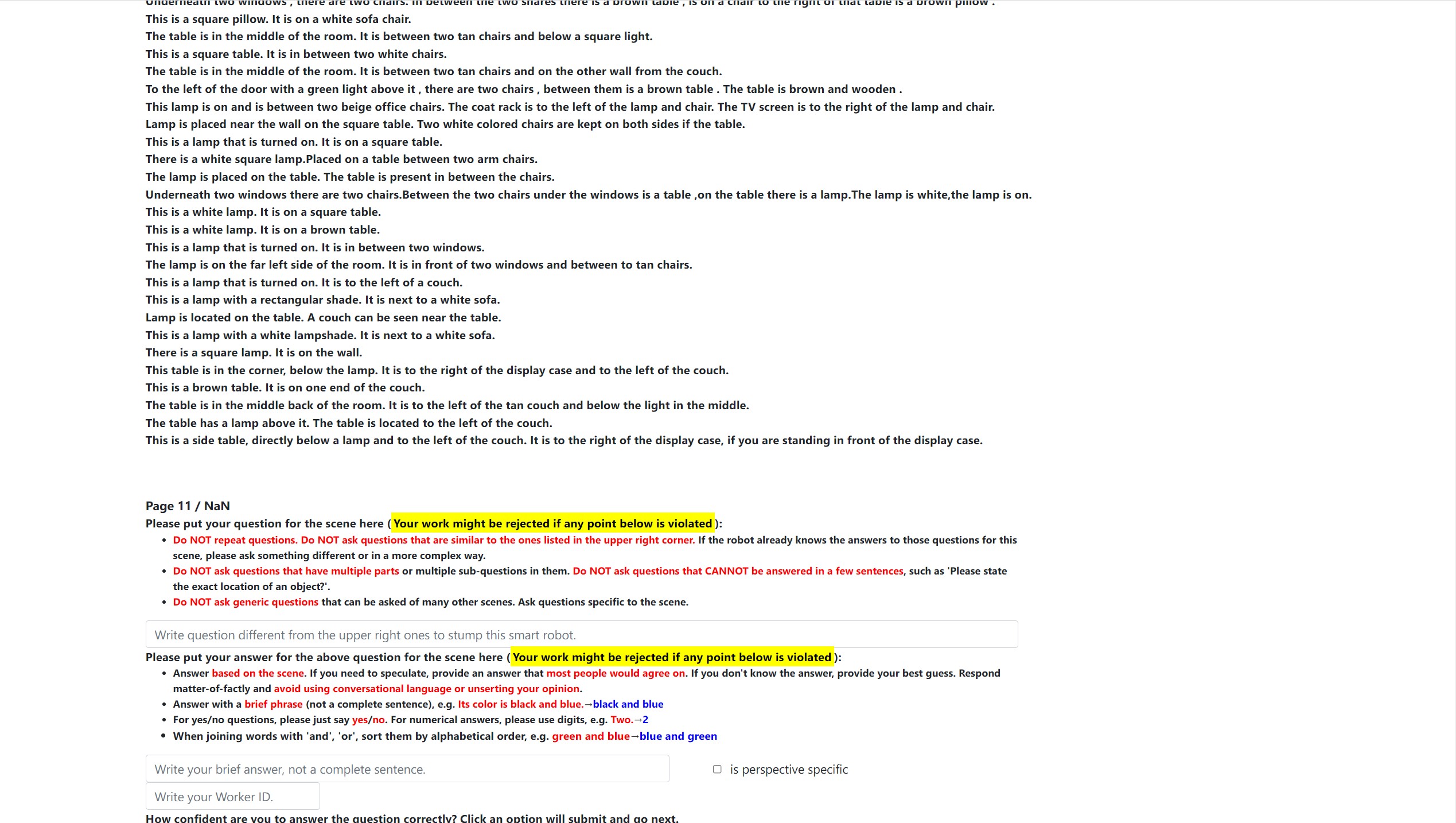}
\caption{Our AMT interface for collecting the answer given captions but without showing frames or scenes.}
\label{supp-webQandcap}
\end{figure*}

\section{Collection Details and Additional Analysis on ScanQA Dataset} 

\subsection{ScanQA Dataset Collection}

\noindent\textbf{Web-based annotation website.} Fig.~\ref{supp-webmain} (left) shows the instruction page of our web-based UI for annotation pipelines of annotation process, as well as basic operations. 
The right image shows several samples to the annotator, which includes some selected interesting questions for different scenes in order to encourage \textit{interesting and diverse questions}. At the request of the annotators, we have made the relevant parts of the price and working hours pixelated.

\subsection{ScanQA Dataset Analysis} \label{subsec:dataanalysis}

\subsubsection{Comparison with VQA}

\noindent\textbf{Question type classification.}
The comparison between ScanQA, VQA and VQA-scene given by Fig. 4 in our main paper classifies the questions based on appearances of different types of words in each question. 
In detail, each question is classified into at most one class among `spatial', `spatial comparison to the average value', `placement', `viewpoint and do navigation', and `aggregation':
\begin{itemize}
\item For `spatial' type of question, it is the question with words `big', `small', `great', `tall', `short', `high', `low', `size', `height', `length', `width', `angle', `parallel', `wide', `vertical', `opposite', `tilt', `diagonal', `near', `by', 
`between', `long', `shape', `next', `fit', `far', `close', `space', `cluttered', `narrow', `spacious', `bigger', `greater', `smaller', `taller', `shorter', `higher', `lower', `longer', `closer', `farther', `inside'.
\item For `spatial comparison to the average value' type of question, it is the question with words `average', `large scene', `small scene', `large sized', `small sized', `high ceiling'.
\item For `placement' type of question, it is the question with words `placed', `arrange', `neat', `perpendicular', `straight', `form', `symmetrical', `covered', `put', `horizontal', `surrounded', `row', `line', `separate', `partition', `sloped'.
\item For `viewpoint and do navigation' type of question, it is the question with words `stand', `sit', `walk', ` go ', `into', `face', `facing', `look', `see', `your', `against', `above', `middle', `side'.
\item For `aggregation' type of question, it is the question with words `many', `all', `any', `largest', `farthest', `most', `least', `nearest', `total', `nothing', `highest', `brightest', `else', `special', ` every', `entire', `totally', `only'.
\end{itemize}

\subsection{Details on Human Accuracy Collection}

In this section, we show the user interface and details for collecting human accuracy that are shown in Table 2 of our main paper.
The left image in Fig.~\ref{supp-webQonlyandQimg} shows the web-based UI for collecting human answers to input setting of `Question' (to answer the question without frames or scenes).
The right image shows the UI for the setting of `Question+Image' (to answer the question with image frame).
For the setting of `Question+Scene', it shares the same UI with our answer collection.
For fairness, the answers are collected by crowdsourcing on AMT, and the tested people are not the annotators for our ScanQA dataset.

We show an additional input setting in Fig.~\ref{supp-webQandcap}, where the tested people are given question and 10 referring expressions of different objects in this scene (Question+Caption). 
The captions are from ScanRefer~\cite{chen2020scanrefer}.
The human accuracy in this setting is shown in Table 2 in our main paper.

\section{Additional Implementation Details}

The size of the candidate answer vocabulary is $166$ for both training and testing.
We train on $4$ 2080Ti GPUs with a total batch size of $64$, using the AdamW optimizer~\cite{loshchilov2018decoupled} and an initial learning rate of $1.0e-8$. The learning rate is updated according to the cyclical learning rate policy.

\section{Additional Limitation}
In addition to the limitation of scene variety as analyzed in our main paper, we encountered another shortcoming of our ScanQA dataset. 
As shown in Fig. 5 in our main paper, among the top questions, the answers for the questions that begin with `can you' are heavily unbalanced - the proportion of `yes' is roughly three times that of `no' and other answers. Phenomenons like this in ScanQA exactly reflect the \textbf{real-world} data bias. However, this makes some questions generally answerable without requiring extensive scene comprehension. Despite carefully designed data collection methods for collecting interesting and diverse questions and high-quality answers, our task is still as likely to suffer from real-world prior, as a slew of widely used VQA datasets~\cite{antol2015vqa,gao2015you,ren2015exploring,yu2015visual,zhang2016yin,tapaswi2016movieqa,zhu2016visual7w,krishna2017visual,johnson2017clevr,agrawal2018don} as studied in ~\cite{agrawal2016analyzing,goyal2017making,agrawal2018don}.%
It may further affect the generalization ability of our framework.
For this issue, a number of efforts have been made in the past few years. 
Goyal et al.\cite{goyal2017making} and Zhang et al.\cite{zhang2016yin} propose new datasets and data collection interfaces based on VQA to balance answer distribution. While the prior is controlled to some extent, the distribution in train and test set are still similar. %
Agrawal et al.\cite{agrawal2018don} propose new splits of VQA to differ the question-answer distribution of train and test sets.
Clevr\cite{johnson2017clevr} generates a diagnostic dataset with little biases to test the model's visual reasoning abilities.
GQA~\cite{hudson2019gqa} extends it to real-world images by providing fine-grained scene graphs and developing an effective question generation method.
Recently, we are witnessing more works on this issue~\cite{amizadeh2020neuro,chen2021meta}. However, they are generally task-and-dataset specific, and it is non-trivial to apply on this 3DQA task.
We are currently investigating data generation with extra annotations to perform systematical analysis and developing models with enhanced robustness to the bias. We believe our work, on the one hand, provides insights into the spatial aspects of the 3DQA task - 3D spatial understanding, navigation, and aggregation; on the other hand, it motivates future study on this challenging issue in this new field.

\end{document}